\documentclass[lettersize,journal]{IEEEtran}
\usepackage{amsmath,amsfonts}
\usepackage{algorithmic}
\usepackage{algorithm}
\usepackage{array}
\usepackage[caption=false,font=normalsize,labelfont=sf,textfont=sf]{subfig}
\usepackage{textcomp}
\usepackage{stfloats}
\usepackage{url}
\usepackage{verbatim}
\usepackage{graphicx}
\usepackage{cite}
\usepackage{multirow}
\usepackage{booktabs}
\begin{document}

\title{Dual Preintegration for Relative State Estimation}

\author{Ruican Xia, Hailong Pei
\thanks{Ruican Xia and Hailong Pei are with the Key Laboratory of Autonomous Systems and Networked Control, Ministry of Education, Unmanned Aerial Vehicle Systems Engineering Technology Research Center of Guangdong, South China University of Technology, Guangzhou, 510640, China.}
\thanks{This work was supported in part by the National Key R\&D Program of China [2023YFB4704900], Aeronautical Science Foundation of China [20220056060001], Scientific Instruments Development Program of NSFC [61527810], New Generation of Information Technology Innovation Project, China University Innovation Fund [2022IT046] and the Fundamental Research Funds for the Central Universities.}}

\markboth{Journal of \LaTeX\ Class Files,~Vol.~14, No.~8, August~2021}%
{Shell \MakeLowercase{\textit{et al.}}: A Sample Article Using IEEEtran.cls for IEEE Journals}

\IEEEpubid{0000--0000/00\$00.00~\copyright~2021 IEEE}

\maketitle

\begin{abstract} 
Relative State Estimation perform mutually localization between two mobile agents undergoing six-degree-of-freedom motion. Based on the principle of circular motion, the estimation accuracy is sensitive to nonlinear rotations of the reference platform, particularly under large inter-platform distances. This phenomenon is even obvious for linearized kinematics, because cumulative linearization errors significantly degrade precision. In virtual reality (VR) applications, this manifests as substantial positional errors in 6-DoF controller tracking during rapid rotations of the head-mounted display. The linearization errors introduce drift in the estimate and render the estimator inconsistent. In the field of odometry, IMU preintegration is proposed as a kinematic observation to enable efficient relinearization, thus mitigate linearized error. Building on this theory, we propose dual preintegration, a novel observation integrating IMU preintegration from both platforms. This method serves as kinematic constraints for consecutive relative state and supports efficient relinearization. We also perform observability analysis of the state and analytically formulate the accordingly null space. Algorithm evaluation encompasses both simulations and real-world experiments. Multiple nonlinear rotations on the reference platform are simulated to compare the precision of the proposed method with that of other state-of-the-art (SOTA) algorithms. The field test compares the proposed method and SOTA algorithms in the application of VR controller tracking from the perspectives of bias observability, nonlinear rotation, and background texture. The results demonstrate that the proposed method is more precise and robust than the SOTA algorithms.

\end{abstract}


\section{Introduction}
\IEEEPARstart{R}{elative} State Estimation (RSE) refers to the process of determining the pose (position and orientation) and velocity of a follower platform relative to a leader platform, where both operate as mobile agents with six-degree-of-freedom (6-DoF) motion. This technique finds critical applications in formation control systems \cite{ref00}, 
aerial refueling tasks \cite{ref01}, and target tracking scenarios \cite{ref02}. The estimations can be classified into direct and indirect methods depending on whether the relative state can be estimated instantaneously \cite{ref37}. The former generally localizes each platform globally at first with odometry, then calculates the relative state from the subtraction between platforms’ global states, while the latter estimate the relative state from direct robot-to-robot observations. This paper focus on the direct method.

Different from conventional localization methods that determine geometric information relative to a fixed reference frame, RSE must account for a dynamically moving reference frame undergoing full 6-DoF motion, where nonlinearity of the 3-DoF rotation introduces significant estimation challenges, since general kinematic formulations often rely on first-order linear approximations. The linearization error becomes particularly pronounced when the inter-platform distance is considerable. Consequently, effective RSE algorithms must explicitly address these nonlinear effects and mitigate cumulative linearization errors in kinematic modeling.

Conventional localization systems employ inertial kinematics that perform state predictions based on linearization around an operating point (derived from prior inertial state). Each acquisition of inertial measurements (angular velocity and linear acceleration) triggers a forward state prediction. However, the inherent presence of higher-order nonlinearities in real-world motion profiles induces progressive error accumulation over successive prediction cycles. While  state update with external measurement can partially mitigate these prediction errors, they fail to achieve complete error elimination. The persistent residual errors become embedded in the system, and the error-containing state is then used as the new operating point for subsequent predictions, leading to further error propagation. Although all filter-based localization schemes exhibit cumulative linearization errors, inertial localization benefits from a fixed reference frame where higher-order terms in the motion model are typically negligible. This results in extremely slow error accumulation, enabling existing frameworks like MSCKF \cite{ref03} to maintain stable performance.

To reduce cumulative linearization errors in large-scale scenarios, an alternative kinematic model known as IMU preintegration simultaneously constrains the previous optimized results and the current state while operating independently of instantaneous state values. The resultant smoother optimizes at least two consecutive states, thereby mitigating error accumulation to some extent. This methodology was originally proposed by T. Lupton and S. Sukkarieh \cite{ref14} to aggregate multiple IMU measurements between keyframes into a single motion constraint. Subsequent advancements include: C. Forster et al.'s \cite{ref15} introduction of SO(3)-based attitude representation, eliminating singularity issues inherent in Euler angle parameterization; T. Qin et al.'s \cite{ref17} derivation of continuous-time IMU preintegration with quaternion-based orientation encoding, particularly suited for computationally constrained platforms like UAVs. Through IMU pre-integration, smoothing-based estimators gain kinematic constraints that permit re-linearization without redundant re-integration computations. This dual capability simultaneously reduces cumulative linearization errors and optimizes computational efficiency.  

The RSE framework establishes its motion model through inertial kinematic principles with state propagation in the reference coordinate system. When IMU measurements from both leader and follower platforms become available, the relative states undergo forward propagation, forming the dynamics for corresponding filtering algorithms. However, this formulation introduces unique challenges distinct from traditional inertial kinematics. Specifically, the rotation of the reference platform introduces the unmeasurable angular acceleration in the relative acceleration. Fosbury et al. \cite{ref1} derived these terms through Euler's equations, a method requiring precise applied torque measurements typically only feasible in space environments. For ground-based implementations, numerical differentiation of angular velocity measurements becomes necessary, introducing susceptibility to nonlinear motion artifacts. In contrast, Xun et al. \cite{ref37} developed an alternative formulation by strategically neglecting rotational-induced linear velocity components. This simplification eliminate angular acceleration computations while maintaining modeling accuracy, thereby enabling stable filter performance. Nevertheless, their approach demonstrates limitations under conditions, where highly nonlinear reference platform rotations induce significant linearization errors since the marginalization of the filter locks in the error. 

The observability characteristics of IMU biases present another critical consideration. The state propagation mechanism employing relative angular velocity and acceleration generates unobservable subspaces for IMU biases during specific relative motions, particularly parallel translations. While such motion patterns rarely persist in practical scenarios, their transient occurrence can trigger bias divergence. Furthermore, approximate realizations of these special motions degrade bias convergence rates, ultimately impacting estimation accuracy. Recent theoretical work by Lai et al. \cite{ref45} systematically investigates these observability constraints. Their analysis establishes that relative rotational motion provides sufficient excitation for complete bias observability, while characterizing the unobservable subspaces under various special motion conditions. This theoretical framework not only explains previous empirical observations but also quantifies the minimum excitation requirements for reliable bias estimation.

Considering the sensitivity of RSE to nonlinear rotations of the reference platform and the capability of IMU preintegration to mitigate accumulated linearization errors, this work leverages IMU preintegration from both leader and follower platforms to establish a kinematic constraint known as dual preintegration for consecutive relative states. The proposed kinematic model employs a simplified relative velocity formulation that preserves model accuracy while avoiding high-order terms. Regarding bias observability, our theoretical analysis adopts a distinct derivation approach, yet aligns with Lai et al.'s conclusion: the dual preintegration framework inherently contains unobservable subspaces for sensor biases during specific motions. We formally prove that relative rotational motion guarantees complete bias observability. Building on this constraint, we develop a maximum a posteriori (MAP) estimator and design a smoothing algorithm. Extensive experiments validate the proposed method through both simulations and real-world data. Simulation studies with varying degrees of nonlinear motion demonstrate that our smoother achieves reduced estimation errors compared to state-of-the-art methods, albeit with increased computational overhead. Furthermore, through simulations of special motion, our experiments demonstrate that the dual preintegration method for IMU bias estimation inherently contains multiple distinct-dimensional unobservable subspaces. For real-world validation, we implement a visual tracking system of a virtual reality (VR) 6-DoF controller using the proposed smoother, along with two benchmark algorithms: an indirect odometry-based method and a direct filter-based RSE approach. Four designed test cases systematically evaluate the algorithm's sensitivity to the biases observability characteristics and the effects of nonlinear rotations of the head-mounted display (HMD). Experimental results confirm that the dual preintegration-based smoother exhibits superior accuracy and enhanced stability.
\begin{figure}[!t]
	\centering
	\includegraphics[width=3.2in]{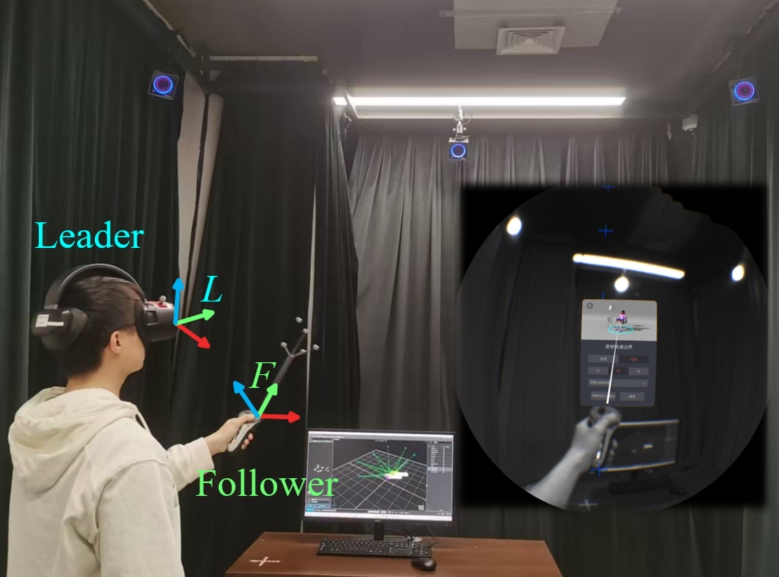}
	\caption{RSE of VR 6-DoF controller.}
	\label{fig_0}
\end{figure}

The remaining sections of this paper are structured as follows. Section \ref{sec2} present the framework overview, where the problem of relative state estimation is formulated as a MAP estimate. Section \ref{sec3} introduces the profile of IMU preintegration, upon which the dual preintegration theory is elaborated in Section \ref{sec4}. The smoother associated to the proposed MAP estimate is validated through simulation and real-world experiments in Section \ref{sec5}. Finally, the conclusions are presented in Section \ref{sec4} followed by the appendix.

\section{Framework Overview}\label{sec2}
\subsection{Problem Formulation}
The RSE problem for leader-follower systems is formulated as a MAP estimation framework. The proposed formulation systematically integrates the novel dual preintegration constraints with multimodal relative measurements, including but not limited to visual correspondence features and inter-agent distance observations. 

The complete state vector at time step $t_i$ comprises:
\begin{equation}
	\mathbf{x}_i \doteq \left\{ \underbrace{\mathrm{R}_F^L, \mathbf{p}_{F\mid L}^{L}, \mathbf{v}_{F \mid L}^{\prime L}}_{\text{relative state }\mathbf{x}_{\mathbf{s}}},\underbrace{\mathbf{b}_{Fg}, \mathbf{b}_{Fa}, \mathbf{b}_{Lg}, \mathbf{b}_{La}}_{\text{IMU biases }\mathbf{x}_{\mathbf{b}}}\right\}_i
\end{equation}
where
\begin{itemize}
	\item $\mathrm{R}_F^L \in SO(3)$: Rotation of the follower's frame $(F)$ with respect to the leader's frame $(L)$;
	\item $\mathbf{p}_{F\mid L}^{L}, \mathbf{v}_{F \mid L}^{\prime L} \in \mathbb{R}^3 $ Position and velocity of the follower relative to the leader, expressed in frame $L$; 
	\item $\mathbf{b}_{g}, \mathbf{b}_{a} \in \mathbb{R}^3$: Gyroscope and accelerometer biases of the leader $(L)$ and the follower $(F)$.
\end{itemize} 
Let \((\cdot)_i\) denote the discrete-time indexed quantity at the \(i^{th}\) sampling instant. The relative velocity in our formulation is defined as
\begin{equation}\label{eq1}
	\mathbf{v}_{F \mid L}^{\prime L}\doteq\mathrm{R}_W^L\left(\mathbf{v}_{F \mid W}^{W}-\mathbf{v}_{L \mid W}^{W}\right),
\end{equation}
where $\mathbf{R}_W^L$ represents the rotation matrix from the gravity-aligned world frame $W$ to the leader's body frame $L$. This definition fundamentally differs from the conventional body-frame derivative
$\mathbf{v}_{F \mid L}^{L}\doteq\frac{\partial \mathbf{p}_{F \mid L}^{L}}{\partial t}$ adopted in \cite{ref1}, with the relationship governed by
\begin{equation}\label{eq0}
	\mathbf{v}_{F \mid L}^{\prime L}=\mathbf{v}_{F \mid L}^{L}+\boldsymbol{\omega}_{L}\times\mathbf{p}_{F\mid L}^{L},
\end{equation}
where $\boldsymbol{\omega}_L \in \mathbb{R}^3$ denotes the leader's body-frame angular velocity. This definition simplifies system realization and observability analysis since the kinematic model are irrelevant to angular acceleration. To simplify symbolic notation, unless otherwise specified, this paper omits subscripts and superscripts associated with the body frame in relative states, i.e. $\mathbf{x}_{\mathbf{s}}\doteq \{\mathrm{R}, \mathbf{p}, \mathbf{v}^{\prime}\}$.

Let $\mathcal{K}_k \doteq \{ t_i \in \mathbb{R}^+ \,|\, 0 \leq t_i \leq t_k \}$ denote the ordered sequence of relative measurement timestamps. Our objective is to derive an optimal smoothing solution through MAP estimation over the complete state history $\mathcal{X}_k \doteq \{ \mathbf{x}_i \,|\, i \in \mathcal{K}_k \}$.
This study focuses on visual features as the relative measurement, while the proposed methodology inherently supports extension to alternative sensing types.

\subsection{Inertial Measurement}\label{sec2_1}
The proposed RSE framework integrates inertial measurements from both platforms, with the measurement models formalized as follows. Let the IMU measurements comprise angular velocity $\boldsymbol{\omega}$ and linear acceleration $\mathbf{a}$, which are subject to sensor biases $\mathbf{b}$ and stochastic noise $\boldsymbol{\eta}$. Following established inertial navigation conventions \cite{ref1}, these measurements are modeled as:
\begin{align}
	\tilde{\boldsymbol{\omega}}_{B}&=\boldsymbol{\omega}_{B}+\mathbf{b}_{Bg}+\boldsymbol{\eta}_{Bg},(B\in\{F,L\}), \label{eq2_}\\
	\tilde{\mathbf{a}}_{B} & =\mathrm{R}_{W}^{B}\left(\mathbf{a}_{B \mid W}^{W}-\mathbf{g}^{W}\right)+\mathbf{b}_{Ba}+\boldsymbol{\eta}_{Ba}
	\nonumber\\
	&
	=\mathbf{a}_{B}+\mathbf{b}_{Ba}+\boldsymbol{\eta}_{Ba},\label{eq1_}
\end{align}
where $\tilde{(\cdot)}$ indicates noise-corrupted measurements, $\mathbf{g}^W \in \mathbb{R}^3$ represents gravitational acceleration in world frame $W$, $\boldsymbol{\eta}_{Bg}$ and $\boldsymbol{\eta}_{Ba}$ are independent identically distributed (i.i.d.) zero-mean Gaussian white noise with $\mathbb{E}[\boldsymbol{\eta}_{Bg}\boldsymbol{\eta}_{Bg}^\top] = \sigma_{Bg}^2\mathbf{I}_3$, $\mathbb{E}[\boldsymbol{\eta}_{Ba}\boldsymbol{\eta}_{Ba}^\top] = \sigma_{Ba}^2\mathbf{I}_3$   

The time-varying IMU biases evolve according to first-order Gauss-Markov dynamics:  
\begin{equation}\label{eq3_}
	\dot{\mathbf{b}}_{Ba}=\boldsymbol{\eta}_{Bba}, ~~ \dot{\mathbf{b}}_{Bg}=\boldsymbol{\eta}_{Bbg},
\end{equation}
where $\boldsymbol{\eta}_{Bba}$ and $\boldsymbol{\eta}_{Bbg}$ represent independent white noise processes with $\mathbb{E}[\boldsymbol{\eta}_{Bba}\boldsymbol{\eta}_{Bba}^\top] = \sigma_{Bba}^2\mathbf{I}_3$ and $\mathbb{E}[\boldsymbol{\eta}_{Bbg}\boldsymbol{\eta}_{Bbg}^\top] = \sigma_{Bbg}^2\mathbf{I}_3$.

\subsection{Visual Measurement}\label{sec2_2}
Relative measurements in a RSE framework generally encode essential spatial information about the relative pose between platforms. This work specifically utilizes visual features detected on the follower's fiducial markers, acquired through imagery captured by the leader's calibrated monocular camera. For a fiducial marker $ l $ with its 3D coordinates defined in the follower's body frame $ F $ as $\mathbf{p}_{l}^{F} \in \mathbb{R}^3$, the measurement model is formulated as:  
\begin{align}\label{eq4_}
	\tilde{\boldsymbol{\rho}}_l & = \pi\left(\mathbf{p}_{l}^{L}\right) + \boldsymbol{\eta}_{\rho}, \nonumber\\
	& = \pi\left(\mathbf{R}_F^L \mathbf{p}_{l}^{F} + \mathbf{p}_{F|L}^L\right) + \boldsymbol{\eta}_{\rho},
\end{align}  
where $\pi(\cdot): \mathbb{R}^3 \rightarrow \mathbb{R}^2$ denotes the perspective projection operator governed by the camera's intrinsic parameters, and $\boldsymbol{\eta}_{\rho} \sim \mathcal{N}(\mathbf{0}, \sigma_{\rho}^2\mathbf{I}_{2})$ represents an i.i.d. zero-mean Gaussian white noise process with covariance \(\sigma_{\rho}^2\mathbf{I}_{2}\).  

\subsection{MAP estimate and Factor Graph}\label{sec2_3}
\begin{figure}[!t]
	\centering
	\includegraphics[width=3.5in]{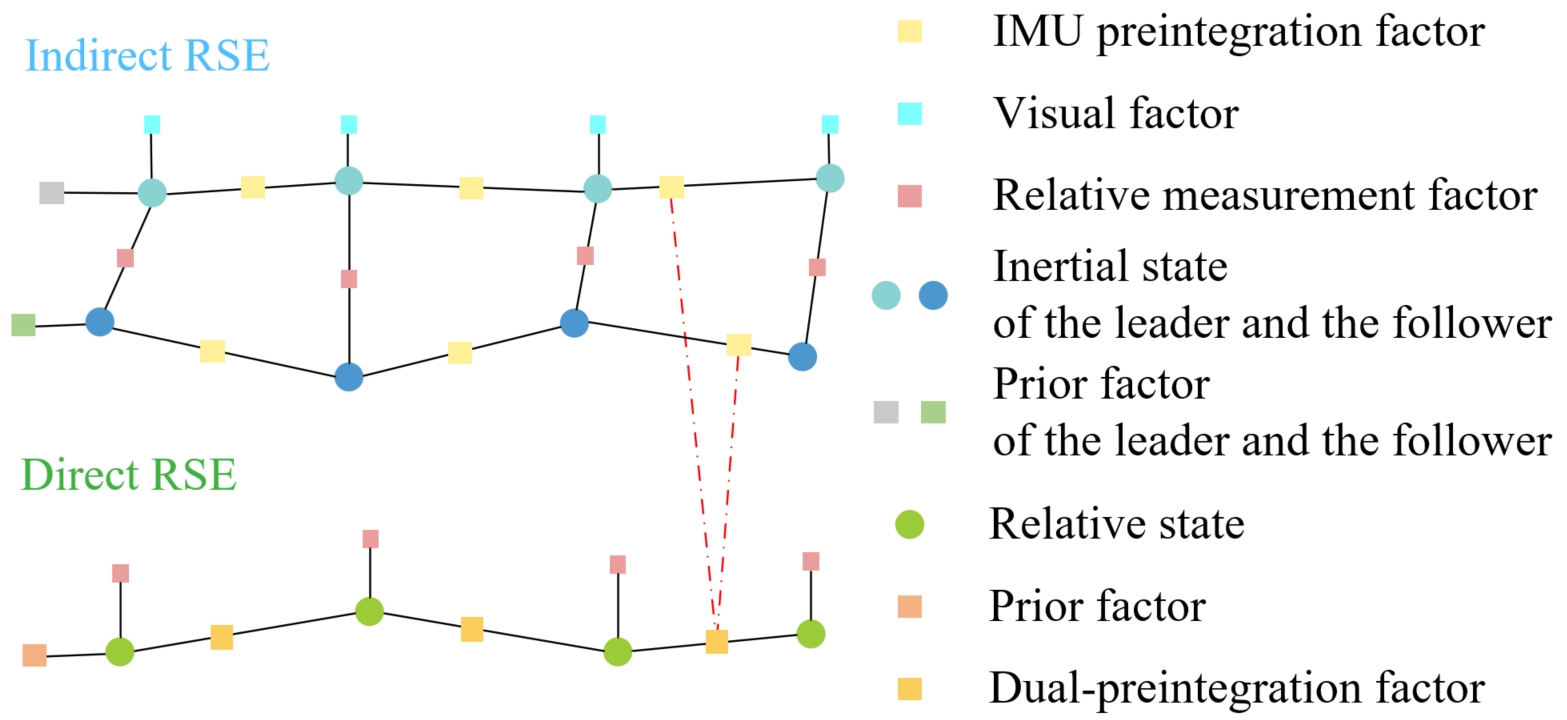}
	\caption{Factor graphs for indirect and direct RSE frameworks. The dual preintegration factor leverages IMU preintegration from both platforms.}
	\label{fig_1}
\end{figure}
The collection of measurements up to timestamp $ k $ is defined as  
\begin{equation}\label{eq5_}  
	\mathcal{Z}_k \doteq \left\{ \mathcal{C}_i, \mathcal{I}_{ij} \right\}_{(i,j)\in\mathcal{K}_k}, \quad \mathcal{I}_{ij} \doteq \left\{ \mathcal{I}_{Fij}, \mathcal{I}_{Lij} \right\}.  
\end{equation}  
where the set $ \mathcal{C}_i $ represents available visual measurements at instant $ i $, comprising multiple feature observations $ \tilde{\boldsymbol{\rho}}_{li} $ ($ l \in \mathcal{C}_i $). Meanwhile, $ \mathcal{I}_{Bij} $ denotes inertial measurements from platform $ B $ between consecutive timestamp $ i $ and $ j $. The posterior probability of states $ \mathcal{X}_k $, conditioned on measurements $ \mathcal{Z}_k $ and priors $ p(\mathcal{X}_0) $, is expressed as:  
\begin{align}\label{eq6_}  
	&p\left( \mathcal{X}_k \mid \mathcal{Z}_k \right)  \propto p\left( \mathcal{X}_0 \right) p\left( \mathcal{Z}_k \mid \mathcal{X}_k \right) \\  
	&= p\left( \mathcal{X}_0 \right) \prod_{(i,j)\in\mathcal{K}_k} p\left( \mathcal{I}_{ij} \mid \mathbf{x}_i, \mathbf{x}_j \right) \prod_{i\in\mathcal{K}_k} \prod_{l\in\mathcal{C}_i} p\left( \tilde{\boldsymbol{\rho}}_{il} \mid \mathbf{x}_i \right), \nonumber  
\end{align}  
implicitly assuming measurement independence and the Markovian property. This probabilistic formulation corresponds to the factor graph in Fig. \ref{fig_1}, where the IMU preintegration from both platforms are unified into the dual preintegration factor, establishing essential relative motion constraints for RSE.  

The MAP estimate \(\mathcal{X}_k^{\star}\) corresponds to the maximizer of Eq. \eqref{eq6_} or equivalently, the minimizer of the negative log-posterior. Under the zero-mean Gaussian noise assumption, this minimization reduces to a nonlinear least squares problem:  
\begin{align}  
	&\mathcal{X}_k^{\star}  \doteq \arg \min _{\mathcal{X}_k} -\log_e p\left(\mathcal{X}_k \mid \mathcal{Z}_k\right)\label{eq7_} \\  
	&=\arg \min _{\mathcal{X}_k} \sum_{(i, j) \in \mathcal{K}_k} \left( \left\|\mathbf{r}_{\mathcal{I}_{ij}} \right\|_{\Sigma_{ij}}^2 + \left\|\mathbf{r}_{\mathbf{b}_{Fij}} \right\|^2 + \left\|\mathbf{r}_{\mathbf{b}_{Lij}} \right\|^2 \right) \nonumber\\  
	&+\sum_{i \in \mathcal{K}_k} \sum_{l \in \mathcal{C}_i} \left\|\mathbf{r}_{\boldsymbol{\rho}_{il}} \right\|_{\Sigma_{\mathcal{C}}}^2 + \left\|\mathbf{r}_0 \right\|_{\Sigma_0}^2, \nonumber  
\end{align}  
where $\mathbf{r}_0$, $\mathbf{r}_{\mathcal{I}_{ij}}$, and $\mathbf{r}_{\boldsymbol{\rho}_{il}}$ denote prior estimate residuals, inertial measurement residuals, and visual measurement residuals, respectively, with corresponding covariance matrices \(\Sigma_0\), \(\Sigma_{ij}\), and \(\Sigma_{\mathcal{C}}\). The bias residual terms are explicitly defined as:  
\begin{equation}  
	\left\|\mathbf{r}_{\mathbf{b}_{Bij}} \right\|^2 \doteq \left\|\mathbf{b}_{Bgj} - \mathbf{b}_{Bgi} \right\|_{\Sigma_{Bg}}^2 + \left\|\mathbf{b}_{Baj} - \mathbf{b}_{Bai} \right\|_{\Sigma_{Ba}}^2,  
\end{equation}  
with covariance matrices $\Sigma_{Bg}$ and $\Sigma_{Ba}$ derived from the bias dynamics in Eq. \eqref{eq3_}. These covariances are computed as $\Sigma_{Bg} = \Delta t_{ij} \sigma_{Bbg}^2 \mathbf{I}_3$ and $\Sigma_{Ba} = \Delta t_{ij} \sigma_{Bba}^2 \mathbf{I}_3$, where $\Delta t_{ij} \doteq \sum_{k=i}^{j-1} \Delta t$ represents the cumulative IMU sampling intervals between timestamp $i$ and $j$.  

The optimal estimate is obtained by solving Eq. \eqref{eq7_} via nonlinear least squares, where each residual term quantifies the discrepancy between sensor measurements and state predictions. This work analytically derives the dual preintegration factor and rigorously establishes the observability of IMU biases under degenerate motion conditions.  

\section{Preliminaries}\label{sec3}
The IMU preintegration process for each platform consolidates sequential inertial measurements into unified motion constraints, explicitly modeling incremental changes in rotation, velocity, and translation. Following the methodology in \cite{ref15}, these constraints are formulated as:  
\begin{align}  
	\Delta \mathrm{R}_{i j} & \doteq \mathrm{R}_{i}^{\top} \mathrm{R}_{j} = \prod_{k=i}^{j-1} \operatorname{Exp} \left( \left( \tilde{\boldsymbol{\omega}}_{k} - \mathbf{b}_{gk} - \boldsymbol{\eta}_{gk} \right) \Delta t \right),  \label{eq4} \\  
	\Delta \mathbf{v}_{i j} & \doteq \mathrm{R}_{i}^{\top} \left( \mathbf{v}_{j} - \mathbf{v}_{i} - \mathbf{g}^W \Delta t_{ij} \right) \label{eq5} \\  
	&= \sum_{k=i}^{j-1} \Delta \mathrm{R}_{i k} \left( \tilde{\mathbf{a}}_{k} - \mathbf{b}_{ak} - \boldsymbol{\eta}_{ak} \right) \Delta t, \nonumber \\  
	\Delta \mathbf{p}_{i j} & \doteq \mathrm{R}_{i}^{\top} \left( \mathbf{p}_{j} - \mathbf{p}_{i} - \mathbf{v}_{i} \Delta t_{ij} - \frac{1}{2} \mathbf{g}^W \Delta t_{ij}^2 \right) \label{eq6} \\  
	&= \sum_{k=i}^{j-1} \left[ \Delta \mathbf{v}_{i k} \Delta t + \frac{1}{2} \Delta \mathrm{R}_{i k} \left( \tilde{\mathbf{a}}_{k} - \mathbf{b}_{ak} - \boldsymbol{\eta}_{ak} \right) \Delta t^2 \right], \nonumber  
\end{align}  
where $ (\cdot)^\top $ denotes matrix transposition and $ \operatorname{Exp}(\boldsymbol{\zeta}) \doteq \exp(\boldsymbol{\zeta}^\wedge) $ represents the exponential map operator, with $ (\cdot)^\wedge $ indicating the skew-symmetric matrix conversion. For notational compactness, we define $ \mathrm{R} \doteq \mathrm{R}_B^W $, $ \mathbf{p} \doteq \mathbf{p}_{B|W}^W $, and $ \mathbf{v} \doteq \mathbf{v}_{B|W}^W $, with platform-specific subscripts omitted due to the generalized applicability across both platforms.  

IMU preintegration inherently combines nominal values with stochastic errors due to sensor noise contamination. To derive covariance propagation, noise terms are decoupled from the right-hand side of Eqs. (\ref{eq4})--(\ref{eq6}) via first-order error approximation:
\begin{align}
	\Delta \mathrm{R}_{i j} &= \prod_{k=i}^{j-1} \operatorname{Exp} \left(\left(\tilde{\boldsymbol{\omega}}_{k}-\mathbf{b}_{gk}\right) \Delta t\right)\operatorname{Exp}\left(-\delta \boldsymbol{\phi}_{i j}\right)
	\nonumber\\&
	=\Delta \tilde{\mathrm{R}}_{i j} \operatorname{Exp}\left(-\delta \boldsymbol{\phi}_{i j}\right),\label{eq_7} \\
	\Delta \mathbf{v}_{i j} &=\sum_{k=i}^{j-1} \Delta \mathrm{R}_{i k}\left(\tilde{\mathbf{a}}_{k}-\mathbf{b}_{ak}\right) \Delta t-\delta \mathbf{v}_{i j}
	\nonumber\\&
	= \Delta \tilde{\mathbf{v}}_{i j}-\delta \mathbf{v}_{i j}, \label{eq_8}\\
	\Delta \mathbf{p}_{i j} &= \sum_{k=i}^{j-1}\left[\Delta \mathbf{v}_{i k} \Delta t+\frac{1}{2} \Delta \mathrm{R}_{i k}\left(\tilde{\mathbf{a}}_{k}-\mathbf{b}_{ak}\right) \Delta t^{2}\right]-\delta \mathbf{p}_{i j}
	\nonumber\\&
	=\Delta \tilde{\mathbf{p}}_{i j}-\delta \mathbf{p}_{i j},\label{eq_9}
\end{align}
where $ \delta \boldsymbol{\phi}_{i j} $, $ \delta \mathbf{v}_{i j} $, and $ \delta \mathbf{p}_{i j} $ denote rotation, velocity, and translation increment errors, respectively. The derivation of \eqref{eq_7} employs the adjoint property $ \mathrm{R} \operatorname{Exp}(\boldsymbol{\phi}) \mathrm{R}^{\top} = \operatorname{Exp}( \mathrm{R} \boldsymbol{\phi})$ and the Baker-Campbell-Hausdorff (BCH) approximation:  
\begin{align}\label{eq10}
&\operatorname{Ln}\left(\operatorname{Exp}\left(\boldsymbol{\phi}_{1}\right) \operatorname{Exp}\left(\boldsymbol{\phi}_{2}\right)\right)\\
&\approx\left\{\begin{array}{l}\mathrm{J}_{r}^{-1}\left(\boldsymbol{\phi}_{1}\right) \boldsymbol{\phi}_{2}+\boldsymbol{\phi}_{1}, \text { for small } \boldsymbol{\phi}_{2}\end{array}\right.\nonumber
\end{align}
where $ \mathrm{J}_{r} $ denotes the right Jacobian with closed-form expression:  
\begin{equation}  
	\mathrm{J}_{r}\left( \boldsymbol{\phi} \right) = \mathrm{I}_3 - \frac{1 - \cos \phi}{\phi^2} \boldsymbol{\phi}^\wedge + \frac{\phi - \sin \phi}{\phi^3} \left( \boldsymbol{\phi}^\wedge \right)^2. \label{eq23_}  
\end{equation}  
and its inverse is   
\begin{align}  
	\mathrm{J}_r^{-1}(\boldsymbol{\phi}) & = \frac{\phi}{2} \cot \frac{\phi}{2} \mathrm{I}_3 + \left( 1 - \frac{\phi}{2} \cot \frac{\phi}{2} \right) \mathbf{a}\mathbf{a}^{\top} + \frac{\phi}{2} \mathbf{a}^\wedge, \label{eq31}  
\end{align}  
with $ \phi = \|\boldsymbol{\phi}\| $ and $ \mathbf{a} = \boldsymbol{\phi}/\phi $.  

The preintegration errors propagate as:  
\begin{align}  
	\delta \boldsymbol{\phi}_{i j} & \simeq \sum_{k=i}^{j-1} \Delta \tilde{\mathrm{R}}_{k+1 j}^{\top} \mathrm{J}_{r}^{k} \boldsymbol{\eta}_{gk} \Delta t, \label{eq12} \\  
	\delta \mathbf{v}_{i j} & \simeq \sum_{k=i}^{j-1} \left[ -\Delta \tilde{\mathrm{R}}_{i k} \left( \tilde{\mathbf{a}}_{k} - \mathbf{b}_{ai} \right)^\wedge \delta \boldsymbol{\phi}_{i k} \Delta t + \Delta \tilde{\mathrm{R}}_{i k} \boldsymbol{\eta}_{ak} \Delta t \right], \nonumber \\  
	\delta \mathbf{p}_{i j} & \simeq \sum_{k=i}^{j-1} \left[ \delta \mathbf{v}_{i k} \Delta t - \frac{1}{2} \Delta \tilde{\mathrm{R}}_{i k} \left( \tilde{\mathbf{a}}_{k} - \mathbf{b}_{ai} \right)^\wedge \delta \boldsymbol{\phi}_{i k} \Delta t^2 \right. \nonumber \\  
	& \quad \left. + \frac{1}{2} \Delta \tilde{\mathrm{R}}_{i k} \boldsymbol{\eta}_{ak} \Delta t^2 \right]. \nonumber  
\end{align}  
where the rotational error term $\delta \boldsymbol{\phi}_{ij}$ arises from gyroscopic noise accumulation, while $\delta \mathbf{v}_{ij}$ and $\delta \mathbf{p}_{ij}$ incorporate both accelerometer noise and error coupling through the cross-product terms. These errors demonstrate zero-mean Gaussian characteristics and the corresponding covariance can be calculated iteratively \cite{ref15}.  

To circumvent computational redundancy in preintegration recalculation during bias updates within the MAP estimation framework, a first-order bias perturbation approximations is employed:  
\begin{align}
\Delta \tilde{\mathrm{R}}_{i j}\left(\mathbf{b}_{gi}\right)&\simeq \Delta \tilde{\mathrm{R}}_{i j}\left(\bar{\mathbf{b}}_{gi}\right)\operatorname{Exp}\left(\frac{\partial \Delta \bar{\mathrm{R}}_{i j}}{\partial \mathbf{b}_{gi}} \delta \mathbf{b}_{gi}\right), \label{eq15}\\
\Delta \tilde{\mathbf{v}}_{i j}\left(\mathbf{b}_{i}\right) &  \simeq \Delta \tilde{\mathbf{v}}_{i j}\left(\bar{\mathbf{b}}_{i}\right)+\frac{\partial \Delta \bar{\mathbf{v}}_{i j}}{\partial \mathbf{b}_{i}} \delta \mathbf{b}_{i}, \label{eq16_}\\
\Delta \tilde{\mathbf{p}}_{i j}\left(\mathbf{b}_{i}\right) & \simeq \Delta \tilde{\mathbf{p}}_{i j}\left(\bar{\mathbf{b}}_{i}\right)+\frac{\partial \Delta \bar{\mathbf{p}}_{i j}}{\partial \mathbf{b}_{i}} \delta \mathbf{b}_{i},\label{eq17_}
\end{align}
where $\mathbf{b}_{i} \doteq \left[ \mathbf{b}_{gi}, \mathbf{b}_{ai} \right] \in \mathbb{R}^6$ denotes the concatenated bias vector, and $\delta \mathbf{b} \doteq \mathbf{b} - \bar{\mathbf{b}}$ represents the bias perturbation from the linearization point $\bar{\mathbf{b}}$. The analytical Jacobian matrices are derived through recursive formulations:  
\begin{align}
	&\frac{\partial  \bar{\mathrm{R}}_{ij}}{\partial \mathbf{b}_{gi}}=-\Sigma_{k=i}^{j-1}\Delta\bar{\mathrm{R}}_{k+1j}^\top\mathrm{J}_r^k\Delta t,\label{eq18_}\\
	&\frac{\partial \Delta \bar{\mathbf{v}}_{ij}}{\partial \mathbf{b}_{ai}}=-\Sigma_{k=i}^{j-1}\Delta\bar{\mathrm{R}}_{ik}\Delta t,\label{eq19}\\
	&\frac{\partial \Delta \bar{\mathbf{v}}_{ij}}{\partial \mathbf{b}_{gi}}=-\Sigma_{k=i}^{j-1}\Delta\bar{\mathrm{R}}_{ik}\left(\tilde{\mathbf{a}}_{k}-\bar{\mathbf{b}}_{ai}\right)^\wedge \frac{\partial  \bar{\mathrm{R}}_{ik}}{\partial \mathbf{b}_{gi}}\Delta t,\label{eq20_}\\
	&\frac{\partial \Delta \bar{\mathbf{p}}_{ij}}{\partial \mathbf{b}_{ai}}=\Sigma_{k=i}^{j-1}\frac{\partial \Delta \bar{\mathbf{v}}_{ij}}{\partial \mathbf{b}_{ai}}\Delta t-\frac{1}{2}\Delta\bar{\mathrm{R}}_{ik}\Delta t^2,\label{eq21_}\\
	&\frac{\partial \Delta \bar{\mathbf{p}}_{ij}}{\partial \mathbf{b}_{gi}}=\Sigma_{k=i}^{j-1}\frac{\partial \Delta \bar{\mathbf{v}}_{ij}}{\partial \mathbf{b}_{gi}}\Delta t-\frac{1}{2}\Delta\bar{\mathrm{R}}_{ik}\left(\tilde{\mathbf{a}}_{k}-\bar{\mathbf{b}}_{ai}\right)^\wedge \frac{\partial  \bar{\mathrm{R}}_{ik}}{\partial \mathbf{b}_{gi}}\Delta t^2\label{eq22}
\end{align}
where $\mathrm{J}_{rk}\doteq\mathrm{J}_{r}\left(\boldsymbol{\tilde{\omega}}_k-\bar{\mathbf{b}}_{gi}\right)$ is the right Jacobian evaluated at the nominal bias. For notational compactness, we define $\Delta\bar{\mathrm{R}}_{ij} \doteq \Delta\tilde{\mathrm{R}}_{ij}(\bar{\mathbf{b}}_i)$.  

The IMU preintegration terms function as factor nodes within the factor graph, enforcing rigid-body kinematic constraints between consecutive platform states in the odometry framework. Building upon this foundation, we systematically integrate leader and follower IMU preintegrations, formulating a novel dual preintegration factor that establishes constrained transformations between successive relative states.  

\section{Dual Preintegration}\label{sec4}
The IMU preintegration theory establishes kinematic constraints between successive odometric states through motion modeling. Building upon this framework, we derive a novel kinematic constraint governing the relative state transformation of collaborative platforms. This constraint originates from paired preintegration measurements obtained through dual-IMU configurations in leader-follower formations \cite{ref45}, henceforth termed dual preintegration. Our analysis demonstrates that the associated error dynamics maintain first-order unbiased white Gaussian characteristics, qualifying dual preintegration as an effective factor for MAP estimation. Notably, the observability properties of relative states exhibit input-dependent characteristics due to inherent nonlinear observations. While observability holds under general motions, unobservable subspaces concerning sensor biases exist for special motions \cite{ref45}. We provide rigorous observability analysis for these distinct scenarios.

The derivation of relative state constraints commences with IMU preintegration-based state propagation. For relative attitude propagation, given the initial relative orientation $\mathrm{R}_{Fi}^{L}$ at instant $i$, the subsequent relative orientation at instant $j$ is computed through:
\begin{align}\label{eq16}
	\mathrm{R}_{j}=&\left(\mathrm{R}_{Lj}\right)^\top\mathrm{R}_{Li}\mathrm{R}_{i}\left(\mathrm{R}_{Fi}\right)^\top\mathrm{R}_{Fj} \\
	=&\Delta {\mathrm{R}_{Li j}^{\top}} \mathrm{R}_{i}\Delta \mathrm{R}_{Fij},\nonumber
\end{align}
where $\Delta\mathrm{R}_{Li j}$ and $\Delta\mathrm{R}_{Fij}$ denote leader and follower attitude preintegrations respectively, derived from Eq. \eqref{eq4}.

Relative position propagation requires coordinate transformation consistency. Following the position preintegration methodology in Eq. \ref{eq6}, we formulate the differential position relationship:
\begin{align}\label{eq18}
	&\mathrm{R}_{i}\Delta \mathbf{p}_{Fi j}-\Delta \mathbf{p}_{Li j}\\=&\Delta \mathrm{R}_{Li j} \mathbf{p}_{j}-\mathbf{p}_{i}-\mathrm{R}_{Wi}^{L} \mathbf{v}_{F \mid Li}^{W} \Delta t_{ij}.\nonumber
\end{align}
Through algebraic manipulation incorporating Eq. \eqref{eq0}, We obtain the relative position propagation as follows:
\begin{align}\label{eq21}
	\mathbf{p}_{j}=&\Delta {\mathrm{R}_{Li j}^\top}(\mathrm{R}_{i} \Delta \mathbf{p}_{Fi j}-\Delta \mathbf{p}_{Li j}+\mathbf{p}_{i}+\mathbf{v}_{i}^{\prime}\Delta t_{ij}).
\end{align}

Velocity propagation follows analogous derivation. Transforming follower velocity preintegrations (Eq. \eqref{eq5}) to the leader frame and differencing yields:
\begin{align}\label{eq17}
	\mathrm{R}_{i} \Delta \mathbf{v}_{Fij}-\Delta \mathbf{v}_{Li j}=&\Delta \mathrm{R}_{Li j} \mathrm{R}_{Wj}^{L} \mathbf{v}_{F \mid L j}^{W}-\mathrm{R}_{W i}^{L} \mathbf{v}_{F \mid Li}^{W}.
\end{align}
Following the relative velocity definition established in Eq. \eqref{eq1}, we isolate the term $\mathbf{v}_{F \mid Li}^{\prime L}$ through algebraic manipulation and the relative velocity propagation is as follows:
\begin{align}\label{eq20}
	\mathbf{v}_{j}^{\prime}=&\Delta {\mathrm{R}_{Li j}^\top}\left(\mathrm{R}_{i} \Delta \mathbf{v}_{Fi j}-\Delta \mathbf{v}_{Li j}+\mathbf{v}_{i}^{\prime}\right).
\end{align}

The state propagation Eqs. \eqref{eq16}, \eqref{eq21}, and \eqref{eq20} characterize relative state evolution driven exclusively by IMU preintegrations of the dual IMUs. As established in Eqs. \eqref{eq_7}-\eqref{eq_9}, these measurements contain inherent errors. To construct optimization factors, we must characterize error propagation dynamics. Our derivation confirms first-order Gaussian error characteristics, enabling effective covariance management in nonlinear optimization frameworks.



\subsection{Dual preintegration measurements and errors}
Building upon the derived propagation, preintegrated measurements of the dual-IMU enable prediction of relative states. 
These synthesized measurements, designated as dual preintegration terms, inherit error characteristics originating from the constituent IMU sensors' stochastic uncertainties. This section presents a rigorous characterization of their error propagation dynamics.

With noise-contaminated preintegration measurements from both IMUs and the derived propagation, measurements of the relative state can be expressed as  
\begin{align}
	\tilde{\mathrm{R}}_{j}\left({\mathbf{x}}_i\right)\doteq&\Delta \tilde{\mathrm{R}}_{Li j}^\top\mathrm{R}_{i} \tilde{\mathrm{R}}_{Fi j},\label{eq23}\\
	\tilde{\mathbf{v}}_{j}^{\prime}\left({\mathbf{x}}_i\right)\doteq&\Delta {\mathrm{R}_{Li j}^\top}\left(\mathrm{R}_{i} \Delta \tilde{\mathbf{v}}_{Fi j}-\Delta \tilde{\mathbf{v}}_{Li j}+\mathbf{v}_{i}^{\prime }\right),\label{eq24}\\
	\tilde{\mathbf{p}}_{j}\left({\mathbf{x}}_i\right)\doteq&\Delta {\tilde{\mathrm{R}}_{Li j}^\top}(\mathrm{R}_{i} \Delta \tilde{\mathbf{p}}_{Fi j}-\Delta \tilde{\mathbf{p}}_{Li j}+\mathbf{p}_{i}+\mathbf{v}_{i}^{\prime}\Delta t_{ij}).\label{eq25_}
\end{align}
The dual preintegration measurement is formally defined as the composite triad:  
\begin{equation}
	\mathcal{D} \doteq \left\{ \tilde{\mathbf{R}}_{FL}^{ij}, \tilde{\mathbf{v}}_{FL}^{\prime L}, \tilde{\mathbf{p}}_{FL}^{L} \right\} \in \mathcal{M} \doteq SO(3) \times \mathbb{R}^3 \times \mathbb{R}^3,
\end{equation}
and the corresponding measurement errors are formulated as
\begin{align}\label{eq26_}
	&\delta \boldsymbol{\alpha}_{ij}\doteq\operatorname{Log}\left(\mathrm{R}_{j}^{\top}\tilde{\mathrm{R}}_{j}\left({\mathbf{x}}_i\right)\right),\\
	&\delta\boldsymbol{\beta}_{ij}\doteq
	\tilde{\mathbf{v}}_{j}\left({\mathbf{x}}_i\right)^{\prime}-
	{\mathbf{v}}_{j}^{\prime},\\
	&\delta\boldsymbol{\gamma}_{ij}\doteq\tilde{\mathbf{p}}_{j}\left({\mathbf{x}}_i\right)-{\mathbf{p}}_{j}.
\end{align}

To derive the error propagation model, we substitute Equations \eqref{eq_7}-\eqref{eq_9} into \eqref{eq16}, \eqref{eq21}, and \eqref{eq20}, retaining only first-order error terms to express the true relative state in \eqref{eq26_} as a composition of preintegrated measurement and error. Through algebraic manipulation, we establish the linear relationship:
\begin{equation}
	\mathbf{w}_{ij}=\Phi \mathbf{n}_{ij}\label{eq25}
\end{equation}
where $\mathbf{w}_{ij}\doteq[\delta \boldsymbol{\alpha}_{ij},\delta\boldsymbol{\beta}_{ij},\delta\boldsymbol{\gamma}_{ij}]\in\mathbb{R}^{9}$ is the vector of dual preintegration error, $\mathbf{n}_{ij}\doteq\left[\mathbf{n}_{Fij}, \mathbf{n}_{Lij}\right]\in\mathbb{R}^{18}$ is the vector of the IMU preintegration errors of both platforms with $\mathbf{n}_{Bij}\doteq[\delta\boldsymbol{\phi}_{Bi j},\delta \mathbf{v}_{Bi j},\delta \mathbf{p}_{Bi j}]\in\mathbb{R}^{9}$,($B\in{F,L}$), and $\Phi\doteq\left[\Phi_{F}, \Phi_{L}\right]$, with
\begin{align}
	&\Phi_{F}=\left[\begin{matrix}
		-\mathrm{I}_3 & \mathrm{0}_3 & \mathrm{0}_3\\
		\mathrm{0}_3 & \Delta \tilde{\mathrm{R}}_{Li j}^\top\mathrm{R}_{i} & \mathrm{0}_3\\
		\mathrm{0}_3 & \mathrm{0}_3 & \Delta \tilde{\mathrm{R}}_{Li j}^\top\mathrm{R}_{i}
	\end{matrix}\right],\\
	&\Phi_{L}=\left[\begin{matrix}
		\tilde{\mathrm{R}}_{j}^{\top} & \mathrm{0}_3 & \mathrm{0}_3 \\
		\left[\tilde{\mathbf{v}}_{j}^{\prime}\times\right] & -\Delta \tilde{\mathrm{R}}_{Li j}^\top & \mathrm{0}_3 \\
		\left[\tilde{\mathbf{p}}_{j}\times\right] & \mathrm{0}_3  & -\Delta \tilde{\mathrm{R}}_{Li j}^\top
	\end{matrix}\right].
\end{align}

The error propagation reveals that the dual preintegration measurement error maintains a linear relationship with the IMU preintegration error and thus preserves first-order zero-mean Gaussian characteristics. These properties enable straightforward covariance computation, which is essential for properly weighting the residuals in the MAP estimator as introduced in Section \ref{Sec4_4}.

\subsection{Measurement update}
Both leader and follower IMU preintegrations depend on accurate sensor bias estimates $\{\mathbf{b}_{Ba}, \mathbf{b}_{Bg}\}\ (B \in \{F,L\})$. During optimization iterations, when bias parameters are updated, we efficiently recompute  through first-order Taylor expansions (Eqs. \eqref{eq15}-\eqref{eq17_}) rather than re-integrating raw measurements. This computational strategy extends to dual preintegration measurements through their dependence on both IMU preintegrations and relative states. 

The dual preintegration update mechanism differs fundamentally from standard IMU preintegration due to its joint dependence on sensor biases and relative states. For a state perturbation $\mathbf{x}_i \leftarrow \bar{\mathbf{x}}_i + \delta\mathbf{x}_i$, the measurement updates follow:
\begin{align}
	&\tilde{\mathrm{R}}_{j}\left({\mathbf{x}}_i\right)\simeq\tilde{\mathrm{R}}_{j}\left(\bar{\mathbf{x}}_i\right)\operatorname{Exp}\left(\frac{\partial \bar{\mathrm{R}}_{j}}{\partial \mathbf{x}_i}\delta{\mathbf{x}}_i\right)
	,\label{eq26}\\
	&\tilde{\mathbf{v}}_{j}^{\prime}\left({\mathbf{x}}_i\right)\simeq\tilde{\mathbf{v}}_{j}^{\prime }\left(\bar{\mathbf{x}}_i\right)+\frac{\partial \bar{\mathbf{v}}_{j}^{\prime}}{\partial {\mathbf{x}}_i}\delta{\mathbf{x}}_i,\label{eq27}\\
	&\tilde{\mathbf{p}}_{j}\left({\mathbf{x}}_i\right)\simeq\tilde{\mathbf{p}}_{j}\left(\bar{\mathbf{x}}_i\right)+\frac{\partial \bar{\mathbf{p}}_{j}}{\partial {\mathbf{x}}_i}\delta{\mathbf{x}}_i.\label{eq28}
\end{align}

The Jacobian matrices 
$\{\frac{\partial \bar{\mathrm{R}}_{j}}{\partial \mathbf{x}_i},\frac{\partial \bar{\mathbf{v}}_{j}^{\prime}}{\partial {\mathbf{x}}_i},\frac{\partial \bar{\mathbf{p}}_{j}}{\partial {\mathbf{x}}_i}\}$
quantify sensitivity relationships between dual preintegration measurements and state variables. The matrices are computed at the linearization point $\bar{\mathbf{x}}_i$ during measurement formation and the analytical derivation are provided in Appendix \ref{app_1}.

\subsection{Dual preintegration factor}\label{Sec4_4}
According to the dual preintegration measurement model in Eq. \eqref{eq25}, where the measurement noise follows a first-order zero-mean Gaussian distribution, we construct corresponding error terms for optimal state estimation within the smoothing framework. The composite residual vector $\mathbf{r}_{\mathbf{s}_{j}}\doteq\left[\mathbf{r}_{\mathrm{R}_{j}},\mathbf{r}_{{\mathbf{v}}_{j}^{\prime}},\mathbf{r}_{{\mathbf{p}}_{j}}\right]\in\mathbb{R}^9$ comprises three geometrically meaningful components:
\begin{align}
	\mathbf{r}_{\mathrm{R}_{j}}\doteq&\operatorname{Log}\left(\mathrm{R}_{j}^{\top}\tilde{\mathrm{R}}_{j}\left(\bar{\mathbf{x}}_i\right)\operatorname{Exp}\left(\frac{\partial \Delta \bar{\mathrm{R}}_{j}}{\partial \mathbf{x}_i}\delta{\mathbf{x}}_i\right)\right)\label{eq29}\\
	\mathbf{r}_{{\mathbf{v}}_{j}^{\prime}}\doteq&\left[\tilde{\mathbf{v}}_{j}^{\prime}\left(\bar{\mathbf{x}}_i\right)+\frac{\partial \Delta \bar{\mathbf{v}}_{j}^{\prime}}{\partial {\mathbf{x}}_i}\delta{\mathbf{x}}_i\right]-{\mathbf{v}}_{j}^{\prime},\label{eq30_}\\
	\mathbf{r}_{{\mathbf{p}}_{j}}\doteq&\left[\tilde{\mathbf{p}}_{j}\left(\bar{\mathbf{x}}_i\right)+\frac{\partial \Delta \bar{\mathbf{p}}_{j}}{\partial {\mathbf{x}}_i}\delta{\mathbf{x}}_i\right]-{\mathbf{p}}_{j}.\label{eq31_}
\end{align}

These residual formulations incorporate the measurement updates from Eqs. \eqref{eq26}-\eqref{eq28}. During numerical optimization iterations, each state correction requires residual reparameterization about the updated operating point. To efficiently solve for the optimal state correction vector $\delta\mathbf{x}_i$, we analytically derive the Jacobian matrices of these residuals through first-order Taylor expansion, with the closed-form expressions provided in Appendix \ref{app_2} for fast iterative computations.

\subsection{Observability of the dual preintegration}\label{sec4_4}
The dual preintegration measurements exhibit nonlinear characteristics as formulated in Eqs. (\ref{eq23})-(\ref{eq25_}). To integrate these measurements into the MAP estimation framework of Eq. (\ref{eq7_}) and compute optimal state estimates, the associated Fisher information matrix must maintain full rank to satisfy local weak observability conditions \cite{ref46}. As established in Eq. (\ref{eq25}), the measurement covariance matrix itself is full-rank. Given that the Fisher information matrix rank is uniquely determined by its Jacobian matrix rank, we analyze the Jacobian structure defined as  
\begin{equation}
	\mathrm{J}_{\text{dpi}} \doteq \left[\mathrm{J}_{\mathbf{s}_i}, \mathrm{J}_{\mathbf{b}_i}, \mathrm{J}_{\mathbf{s}_j}, \mathrm{J}_{\mathbf{b}_j}\right].
\end{equation}
From the Jacobian derivation in Appendix \ref{app_2}, the submatrices $\mathrm{J}_{\mathbf{s}_i}$ and $\mathrm{J}_{\mathbf{s}_j}$ are structurally block-diagonalizable with full-rank diagonal blocks. The measurement independence from IMU biases at time $j$ renders $\mathrm{J}_{\mathbf{b}_j}$ a null matrix. The critical analysis therefore focuses on $\mathrm{J}_{\mathbf{b}_i}$, whose rank varies with the input characteristics of IMU preintegration increments. To systematically investigate $\mathrm{J}_{\mathbf{b}_i}$'s rank properties under different motion inputs, we formulate the matrix as:
\begin{equation}\label{eq32}
	\mathrm{J}_{\mathbf{b}_i}\doteq\left[\mathrm{J}_{\mathrm{R}_j\mathbf{b}_i},\mathrm{J}_{\mathbf{v}_j^\prime\mathbf{b}_i},\mathrm{J}_{\mathbf{p}_j\mathbf{b}_i}\right],
\end{equation}
where
\begin{align*}
	&\mathrm{J}_{\mathrm{R}_j\mathbf{b}_i}=\mathrm{J}_r^{-1}\left(\bar{\mathbf{r}}_{{{\mathrm{R}_{j}}}}\right)\left[\begin{matrix}
		\frac{\partial \Delta \bar{\mathrm{R}}_{Fij}}{\partial \mathbf{b}_{Fgi}} & \mathrm{0}_3 & -\bar{\mathrm{R}}_{j}^\top\frac{\partial \Delta \bar{\mathrm{R}}_{Lij}}{\partial \mathbf{b}_{Lgi}}& \mathrm{0}_3   
	\end{matrix}\right],\\
	&\mathrm{J}_{\mathbf{v}_j^\prime\mathbf{b}_i}=\left[\begin{matrix}
		\frac{\partial  \bar{\mathbf{v}}_{j}^{\prime}}{\partial \mathbf{b}_{Fgi}} & \frac{\partial  \bar{\mathbf{v}}_{j}^{\prime}}{\partial \mathbf{b}_{Fai}} & \frac{\partial  \bar{\mathbf{v}}_{j}^{\prime}}{\partial \mathbf{b}_{Lgi}} & \frac{\partial  \bar{\mathbf{v}}_{j}^{\prime}}{\partial \mathbf{b}_{Lai}}
	\end{matrix}\right],\\
	&\mathrm{J}_{\mathbf{p}_j\mathbf{b}_i}=\left[\begin{matrix}
		\frac{\partial  \bar{\mathbf{p}}_{j}}{\partial \mathbf{b}_{Fgi}} & \frac{\partial  \bar{\mathbf{p}}_{j}}{\partial \mathbf{b}_{Fai}} & \frac{\partial  \bar{\mathbf{p}}_{j}}{\partial \mathbf{b}_{Lgi}} & \frac{\partial  \bar{\mathbf{p}}{j}}{\partial \mathbf{b}{Lai}}
	\end{matrix}\right],
\end{align*}
The partial derivatives of IMU preintegration with respect to sensor biases are derived in Eqs. \eqref{eq18_}-\eqref{eq22}. Under general relative motion conditions, where leader and follower platforms undergo arbitrary 3D rotations and translations as defined in \cite{ref45}, the Jacobian submatrix $\mathrm{J}_{\mathbf{b}_i}$ maintains full row rank. 

However, practical systems often exhibit motion constraints. A representative case occurs in aerial refueling missions, where the leader and follower platforms maintain relative stationarity after successful docking. This operational requirement introduces additional kinematic constraints that reduce the system's observable degrees of freedom. Under such constrained motion patterns, the Jacobian submatrix $\mathrm{J}_{\mathbf{b}_i}$ may become rank-deficient. Through null space analysis of the Jacobian matrix, we formally characterize these unobservable subspaces. Following the observability analysis framework established in \cite{ref47,ref48}, the unobservable subspace corresponds to the solution set:  
\begin{equation}
	\mathrm{J}_{\mathbf{b}_i}\delta\mathbf{x}_{\mathbf{b}_i} = \mathbf{0}_9
\end{equation}  
where $\delta\mathbf{x}_{\mathbf{b}_i} \in \mathbb{R}^{12}$ denotes the bias perturbation vector defined in Eq. (\ref{eq61}). We formulate the complete union of unobservable subspaces as follows:
\begin{equation}
	\left[\begin{matrix}\label{eq35}
		\mathbf{b}_{g_i}^+ \\ \mathbf{b}_{a_i}^+
	\end{matrix}\right]=
	\left[\begin{matrix}
		\mathrm{R}_{i} & \mathrm{0}_3 & \mathrm{I}_3 & \mathrm{0}_3\\
		\mathrm{0}_3 & \mathrm{R}_{i} & \mathrm{0}_3 & \mathrm{I}_3
	\end{matrix}
	\right].
\end{equation}
The physical meaning of each unobservable subspace is
\begin{itemize}
	\item $\mathbf{b}_{a+}$ is 3 DoF of the composite accelerometer bias.
	\item $\mathbf{b}_{a+}^{\mathbf{\omega}}=\mathbf{b}_{a+}\boldsymbol{\omega}_F^L$ is 1 DoF of the composite gyroscope bias along $\boldsymbol{\omega}_F^L$ direction, where $\boldsymbol{\omega}_F^L\doteq\mathrm{R}\boldsymbol{\omega}_F - \boldsymbol{\omega}_L$ is the relative angular velocity.
	\item $\mathbf{b}_{g+}$ is 3 DoF of the composite gyroscope bias.
	\item $\mathbf{b}_{g+}^{\boldsymbol{\omega}}=\mathbf{b}_{g+}\boldsymbol{\omega}_F^L$ is 1 DoF of the composite gyroscope bias along $\boldsymbol{\omega}_F^L$ direction.
\end{itemize}
Unobservable subspaces induced by specific motion constraints are formally classified as follows:
\begin{enumerate}
	\item if $\boldsymbol{\omega}_F^L\equiv\mathbf{0}_3$, given $\exists \delta\bar{\mathbf{x}}_{\mathbf{b}_i} $ s.t.  $\delta\bar{\mathbf{x}}_{\mathbf{b}_i} = \arg\min_{\delta\mathbf{x}_{\mathbf{b}_i}} \|\mathbf{r}_{\mathbf{s}_{j}}(\delta\mathbf{x}_{\mathbf{b}_i})\|^2$, then for $\forall \delta\mathbf{z} \in \mathbb{R}^3$ and $\delta\mathbf{x}_{\mathbf{b}_i}^* = \delta\bar{\mathbf{x}}_{\mathbf{b}_i} + \mathbf{b}_{a_i}^+ \delta\mathbf{z}$, there exists $\mathbf{r}_{\mathbf{s}_{j}}(\delta\mathbf{x}_{\mathbf{b}_i}^*) = \mathbf{r}_{\mathbf{s}_{j}}(\delta\bar{\mathbf{x}}_{\mathbf{b}_i})$, i.e. $\mathbf{b}_{a_i}^+$ spans an unobservable subspace, which induces a three-dimensional ambiguity in the accelerometer bias estimation for both the leader and follower platforms;
	
	\item if $\boldsymbol{\omega}_{Fk}^L=s_k\boldsymbol{\tau}$, given $\exists \delta\bar{\mathbf{x}}_{\mathbf{b}_i} $ s.t.  $\delta\bar{\mathbf{x}}_{\mathbf{b}_i} = \arg\min_{\delta\mathbf{x}_{\mathbf{b}_i}} \|\mathbf{r}_{\mathbf{s}_{j}}(\delta\mathbf{x}_{\mathbf{b}_i})\|^2$, then for $\forall \delta{z} \in \mathbb{R}$ and $\delta\mathbf{x}_{\mathbf{b}_i}^* = \delta\bar{\mathbf{x}}_{\mathbf{b}_i} + \mathbf{b}_{a_i}^+\boldsymbol{\omega}_F^L \delta{z}$, there exists $\mathbf{r}_{\mathbf{s}_{j}}(\delta\mathbf{x}_{\mathbf{b}_i}^*) = \mathbf{r}_{\mathbf{s}_{j}}(\delta\bar{\mathbf{x}}_{\mathbf{b}_i})$, i.e. $\mathbf{b}_{a_i}^+\boldsymbol{\omega}_F^L$ spans an unobservable subspace, which induces a one-dimensional ambiguity in the accelerometer bias estimation for both the leader and follower platforms;
	
	\item if $\boldsymbol{\omega}_F^L\equiv\mathbf{0}_3$, and simultaneously have: 1. $\mathbf{p}\equiv\mathbf{0}_3$ or $\boldsymbol{\omega}_L\equiv\mathbf{0}_3$, 2. $\mathbf{v}\doteq\mathbf{v}_{F \mid L}^{L}\equiv\mathbf{0}_3$, given $\exists \delta\bar{\mathbf{x}}_{\mathbf{b}_i} $ s.t.  $\delta\bar{\mathbf{x}}_{\mathbf{b}_i} = \arg\min_{\delta\mathbf{x}_{\mathbf{b}_i}} \|\mathbf{r}_{\mathbf{s}_{j}}(\delta\mathbf{x}_{\mathbf{b}_i})\|^2$, then for $\forall \delta\mathbf{z} \in \mathbb{R}^3$ and $\delta\mathbf{x}_{\mathbf{b}_i}^* = \delta\bar{\mathbf{x}}_{\mathbf{b}_i} + \mathbf{b}_{g_i}^+ \delta\mathbf{z}$, there exists $\mathbf{r}_{\mathbf{s}_{j}}(\delta\mathbf{x}_{\mathbf{b}_i}^*) = \mathbf{r}_{\mathbf{s}_{j}}(\delta\bar{\mathbf{x}}_{\mathbf{b}_i})$, i.e. $\mathbf{b}_{g_i}^+$ spans an unobservable subspace, which induces a three-dimensional ambiguity in the gyroscope bias estimation for both the leader and follower platforms;
	
	\item if $\boldsymbol{\omega}_{Fk}^L=s_k\boldsymbol{\tau}$, and simultaneously have: 1. $\mathbf{p}\|\boldsymbol{\omega}_L$ or $\mathbf{p}\equiv\mathbf{0}_3$ or $\boldsymbol{\omega}_L\equiv\mathbf{0}_3$,
	2. $\mathbf{p}\|\boldsymbol{\omega}_F^L$ or $\mathbf{p}\equiv\mathbf{0}_3$ or $\boldsymbol{\omega}_L\equiv\mathbf{0}_3$ and 3. $\mathbf{v}\|\boldsymbol{\omega}_F^L$ or $\mathbf{v}\equiv\mathbf{0}_3$, given $\exists \delta\bar{\mathbf{x}}_{\mathbf{b}_i} $ s.t.  $\delta\bar{\mathbf{x}}_{\mathbf{b}_i} = \arg\min_{\delta\mathbf{x}_{\mathbf{b}_i}} \|\mathbf{r}_{\mathbf{s}_{j}}(\delta\mathbf{x}_{\mathbf{b}_i})\|^2$, then for $\forall \delta{z} \in \mathbb{R}$ and $\delta\mathbf{x}_{\mathbf{b}_i}^* = \delta\bar{\mathbf{x}}_{\mathbf{b}_i} + \mathbf{b}_{g_i}^+\boldsymbol{\omega}_F^L \delta{z}$, there exists $\mathbf{r}_{\mathbf{s}_{j}}(\delta\mathbf{x}_{\mathbf{b}_i}^*) = \mathbf{r}_{\mathbf{s}_{j}}(\delta\bar{\mathbf{x}}_{\mathbf{b}_i})$, i.e. $\mathbf{b}_{g_i}^+\boldsymbol{\omega}_F^L$ spans an unobservable subspace, which induces a one-dimensional ambiguity in the gyroscope bias estimation for both the leader and follower platforms;
\end{enumerate}
Although the derivation methodology in Appendix \ref{app_3} diverges fundamentally from the approach in \cite{ref45}, both analyses conclusively demonstrate identical unobservable subspaces for sensor biases under constrained motion conditions.

\section{Experiment Verification}\label{sec5} 
In this section, we conduct two experiments to validate the proposed MAP estimator on both simulated and real data. Section \ref{sec5_1} compares the proposed algorithm with state-of-the-art methods using synthetic data in MATLAB. Section \ref{sec5_2} evaluates the algorithm's observability under special motion conditions. Section \ref{sec5_3} implements the algorithm for visual tracking of the 6-DoF controller in a commercial VR device, comparing its performance with the manufacturer's proprietary solution and the ground truth. The results demonstrate the superior precision and robustness of the proposed relative navigation framework. The code is publicly available on GitHub: \url{https://github.com/richardXia7462/RelativeStateEstimation}.

\subsection{Algorithm Comparison}\label{sec5_1}
This section compares the pose estimation accuracy between the dual preintegration-based smoother and state-of-the-art relative state estimation algorithms through simulated general motion on MATLAB, focusing on their performance under motions with different nonlinearity levels. The algorithms employ distinct kinematic models, thus have varying accumulated linearization errors. Here, general motion refers to relative rotations and translations along all axes between platforms \cite{ref45}, while motion nonlinearity level is defined by the proportion of higher-order terms in the rotational increments of the reference platform. The study aims to evaluate these algorithms' precision when handling abrupt motion changes, such as back-and-forth rotations in head-mounted devices.

The evaluated algorithms comprise:  
\begin{itemize}
	\item DPFLS: Fixed-lag smoother implementing the proposed MAP estimation framework. This method employs dual preintegration to establish constraints between consecutive states (window size = 2).
	\item DIEKF \cite{ref45}: Dual IMU-driven Extended Kalman Filter using the conventional relative velocity $\mathbf{v}_{F \mid L}^{L}$ from $\eqref{eq0}$. Its kinematic model incorporates angular acceleration components as formulated in Equation $\eqref{eq76}$. 
	\item SESKF \cite{ref37}: Error state Kalman filter adopting the simplified velocity formulation $\mathbf{v}_{F \mid L}^{\prime L}$ from $\eqref{eq0}$ (consistent with DPFLS), while eliminating angular acceleration dependencies.
\end{itemize}  
It should be noted that DIEKF relies on a higher-order system implementation. However, due to the general inaccessibility of angular acceleration in practical  applications, the method employs a first-order linear approximation. To ensure a fair comparative analysis, since all  baseline methods included in the study are  filters, DPFLS is constrained to  execute only a single optimization step.

The simulation trajectory emulates the leader-follower visual tracking routine motion, where the leader platform rotates about a fixed axis  and the follower platform freely rotates. Besides both platforms perform relative translational motion along each axis. Three angular velocity profiles $\omega(t)$ for the leader platform are designed: 
\begin{itemize}
	\item Constant rotation: $\omega_1(t) \equiv \pi$ 
	\item Harmonic motion: $\omega_2(t) = 2\pi\sin(2\pi t)$ 
	\item Stochastic excitation: $\omega_3(t) \sim \mathcal{N}(0,1)$  
\end{itemize} 

According to the definition of the nonlinearity levels, the parameter can be formulated as: $\lambda\doteq\frac{\dot\omega dt}{2\omega}$, where $dt$ represents the sampling interval of the IMUs. These configurations implement progressively increasing nonlinearity levels to evaluate the algorithm's sensitivity to linearization errors. The simulated measurements comprise:
\begin{itemize} 
\item Visual data: 25Hz image acquisition with 1 pixel feature point noise; 
\item Inertial data: 250Hz angular velocity and linear acceleration measurements with noise parameters specified in Table \ref{tab1}.  
\end{itemize} 
\begin{table}[hbt!]
	\caption{IMU continuous-time noise parameters}
	\label{tab1}
	\centering
	\begin{tabular}{cc}
		\toprule
		$\sigma_{g v}[\mathrm{rad} /(s \sqrt{\mathrm{Hz}})]$ & $1.528 \times 10^{-3}$ \\
		$\sigma_{g u}[\mathrm{rad} /(s^{2} \sqrt{\mathrm{Hz}})]$ & $1.867 \times 10^{-4}$ \\
		$\sigma_{a v}[\mathrm{m} /(s^{2} \sqrt{\mathrm{Hz}})]$ & $1.244 \times 10^{-2}$ \\
		$\sigma_{a u}[\mathrm{m} /(s^{3} \sqrt{\mathrm{Hz}})]$ & $7.841 \times 10^{-3}$ \\
		\bottomrule
	\end{tabular}
\end{table}

\begin{figure*}[!t]
	\centering
	\includegraphics[width=6.0in]{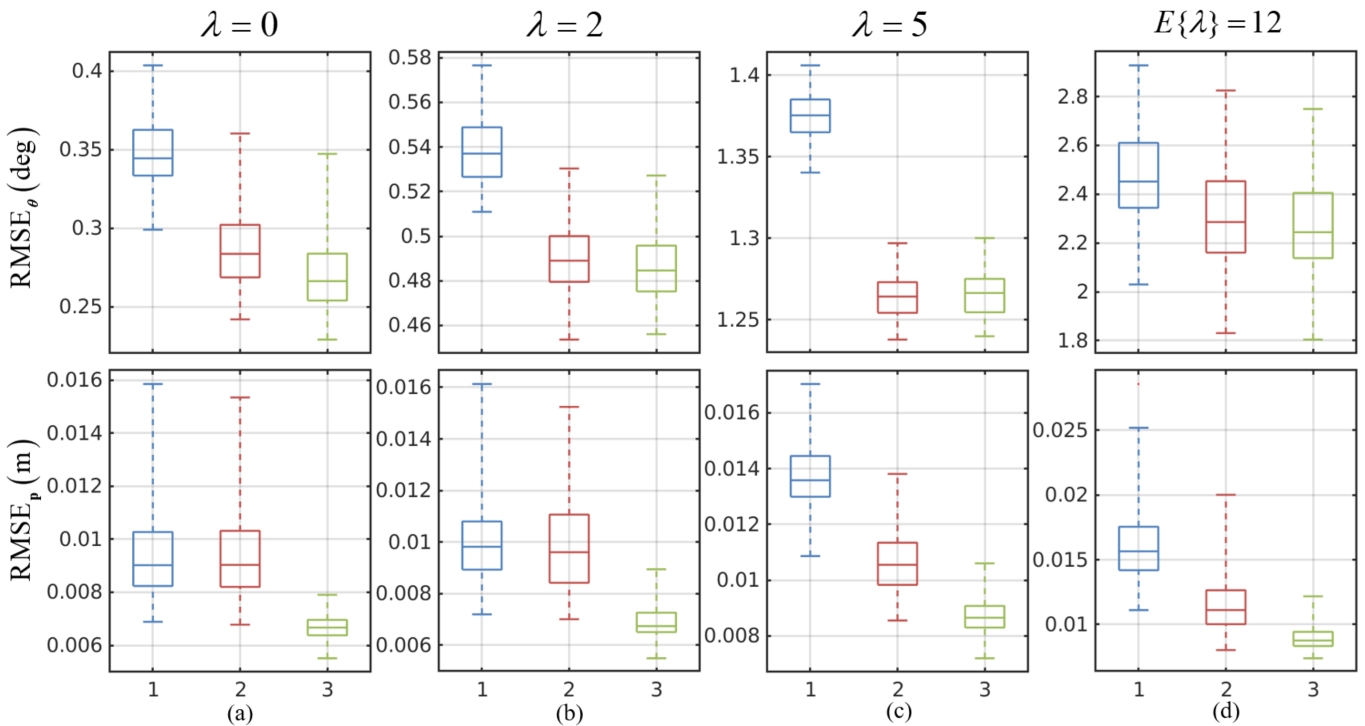}
	\caption{Comparison of algorithms DIEKF (1), SESKF (2) and DPFLS (3) in terms of RMSE of the relative pose with multiple nolinear motion . }
	\label{fig_5}
\end{figure*}

Each algorithm conducts 100 Monte Carlo experiments, and their performance is evaluated based on the root-mean-square errors (RMSEs) of relative rotation and position, denoted as $\textrm{RMSE}_{\boldsymbol{\theta}}$ and $\textrm{RMSE}_{\mathbf{p}}$. Error distributions are statistically summarized in Fig. \ref{fig_5}, with subfigures (a), (b), and (d) depicting the pose estimation error characteristics under the three angular velocity regimes $\omega(t)$, respectively. Furthermore, analytical derivations of $\lambda$ and $\omega_2(t)$ reveal that enlarging $dt$ systematically elevates motion nonlinearity. Thus, we specifically configure the IMU sampling frequency at 100 Hz during $\omega_2(t)$ testing, with the corresponding error distribution statistics presented in subfigure (c). 

When operating at constant rotational velocity $\omega_1(t)$, all motion models demonstrate kinematic consistency with ground truth, yielding minimal pose estimation discrepancies. The $\textrm{RMSE}_{\mathbf{p}}$ and $\textrm{RMSE}_{\boldsymbol{\theta}}$ of all algorithms achieve mean values below 1 cm and 0.5°, respectively. Notably, DIEKF exhibits marginally elevated errors due to its numerical differentiation approach for angular acceleration computation in the kinematic model.  The nonlinear components introduced by the follower platform's free rotation necessitate state window-based reoptimization in DPFLS, enabling a 25.56\% reduction in $\textrm{RMSE}_{\mathbf{p}}$ compared to SESKF. When switching to $\omega_2(t)$ (where $\lambda$ increases from 0 to 2), all algorithms demonstrate approximately 0.2° growth in relative orientation error while maintaining millimeter-level positional accuracy. After extending the IMU sampling interval, the nonlinearity levels reach 5. consequently,  DIEKF's orientation error surges by 56\% (0.84°) and SESKF and DPFLS degrade to 1.26°.  Notably, while DIEKF's $\textrm{RMSE}_{\mathbf{p}}$ reaches 1.36 cm, SESKF preserves more accurate localization through precise motion modeling despite its filter-based state estimation. DPFLS consistently outperforms others in positional accuracy. Under stochastic Gaussian-distributed angular velocity (mean $\lambda=12$), orientation errors increase by >1° across all methods and positional errors escalate to 1.56 cm (DIEKF), 1.11 cm (SESKF), and 0.87 cm (DPFLS). The enhanced accuracy of DPFLS incurs a 178\% time penalty (2.5 ms/update vs. 0.9 ms for EKF variants) on an Intel i9-10900K platform, attributable to doubled state dimensionality. 

The proposed smoother-based approach, leveraging dual preintegration for consecutive state constraints, effectively mitigates linearization errors in RSE despite increased computational load. This demonstrates a viable paradigm for motion-constrained estimation systems prioritizing precision over real-time efficiency.

\subsection{Observability Eveluation}\label{sec5_3}

In the preceding section, trajectories representing general motion were employed to conduct an accuracy comparison among estimation algorithms grounded in different relative motion models. This analysis demonstrated that the dual preintegration-based smoother enhances estimation accuracy via simultaneously estimate a set of recent states with motion constraint and relinearize past measurements. Nevertheless, the method fails to surmount the observability issue of the biases. In this section, experiments are carried out to verify that this phenomenon aligns with both the theoretical derivation outcomes of this paper and the conclusions put forward in \cite{ref45}. 

Analogous to the experimental protocol in \cite{ref45}, for a variety of special motions, the error and its $3\sigma$ criterion are utilized to assess the accuracy and consistency of the estimation results. This assessment, in turn, offers insights into the observability of the biases. 

\begin{figure}[htpb!]
	\centering
	\includegraphics[width=3.4in]{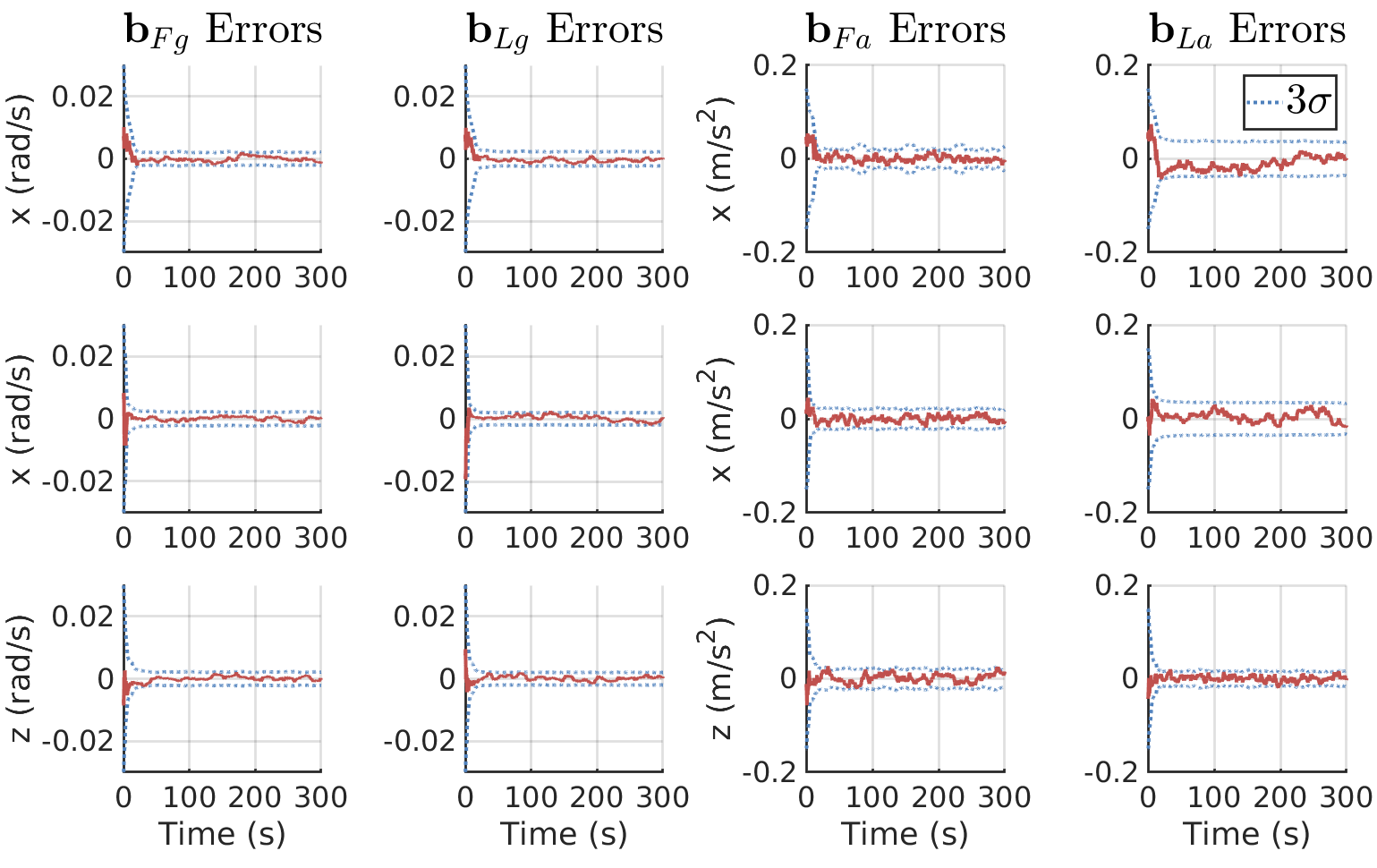}
	\caption{Convergence characteristics of estimation biases for case 1.}\label{figa}
\end{figure}

\begin{figure}[htpb!]
	\centering
	\includegraphics[width=3.4in]{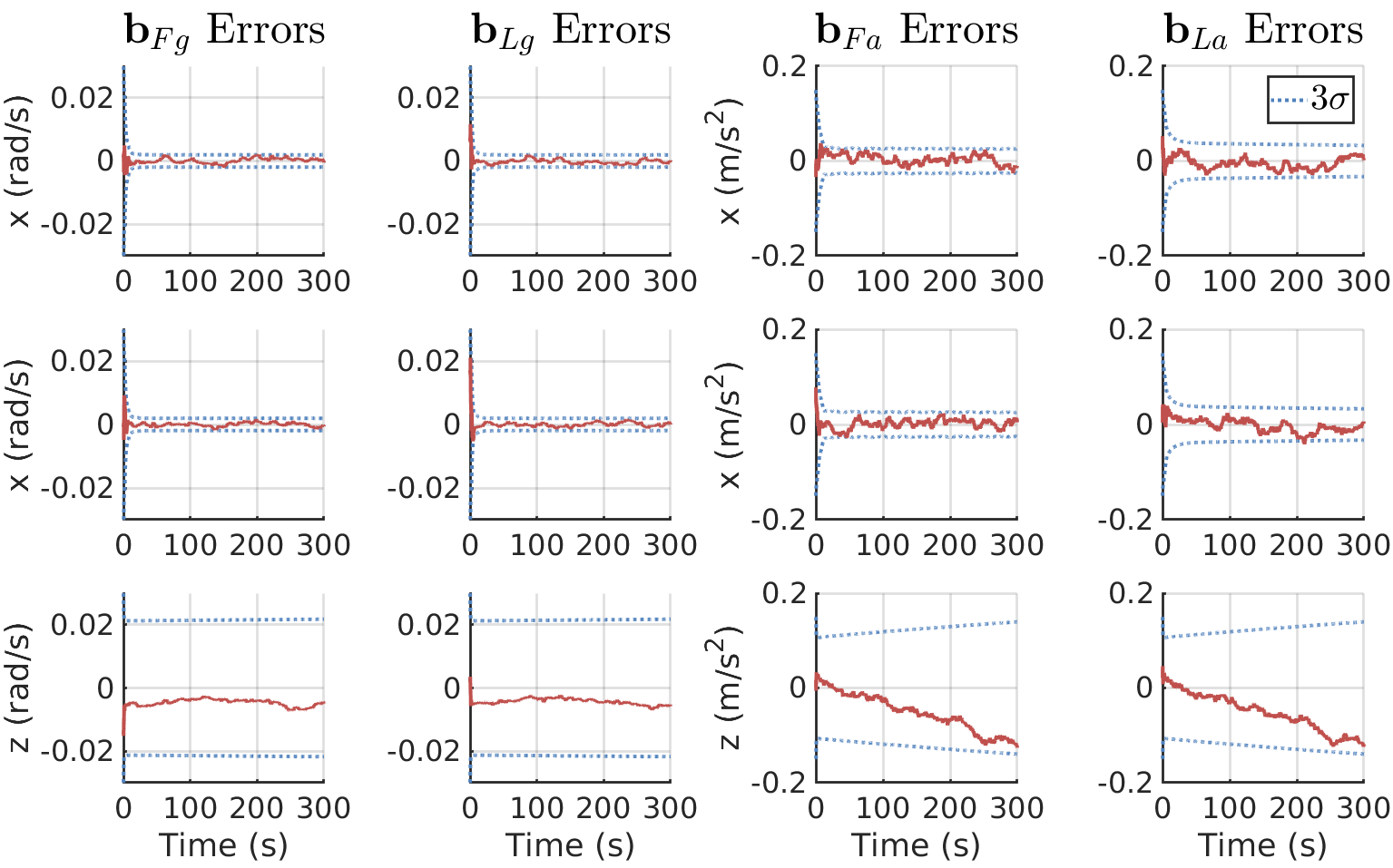}
	\caption{Convergence characteristics of estimation biases for case 2.}\label{figb}
\end{figure}

\begin{figure}[htpb!]
	\centering
	\includegraphics[width=3.4in]{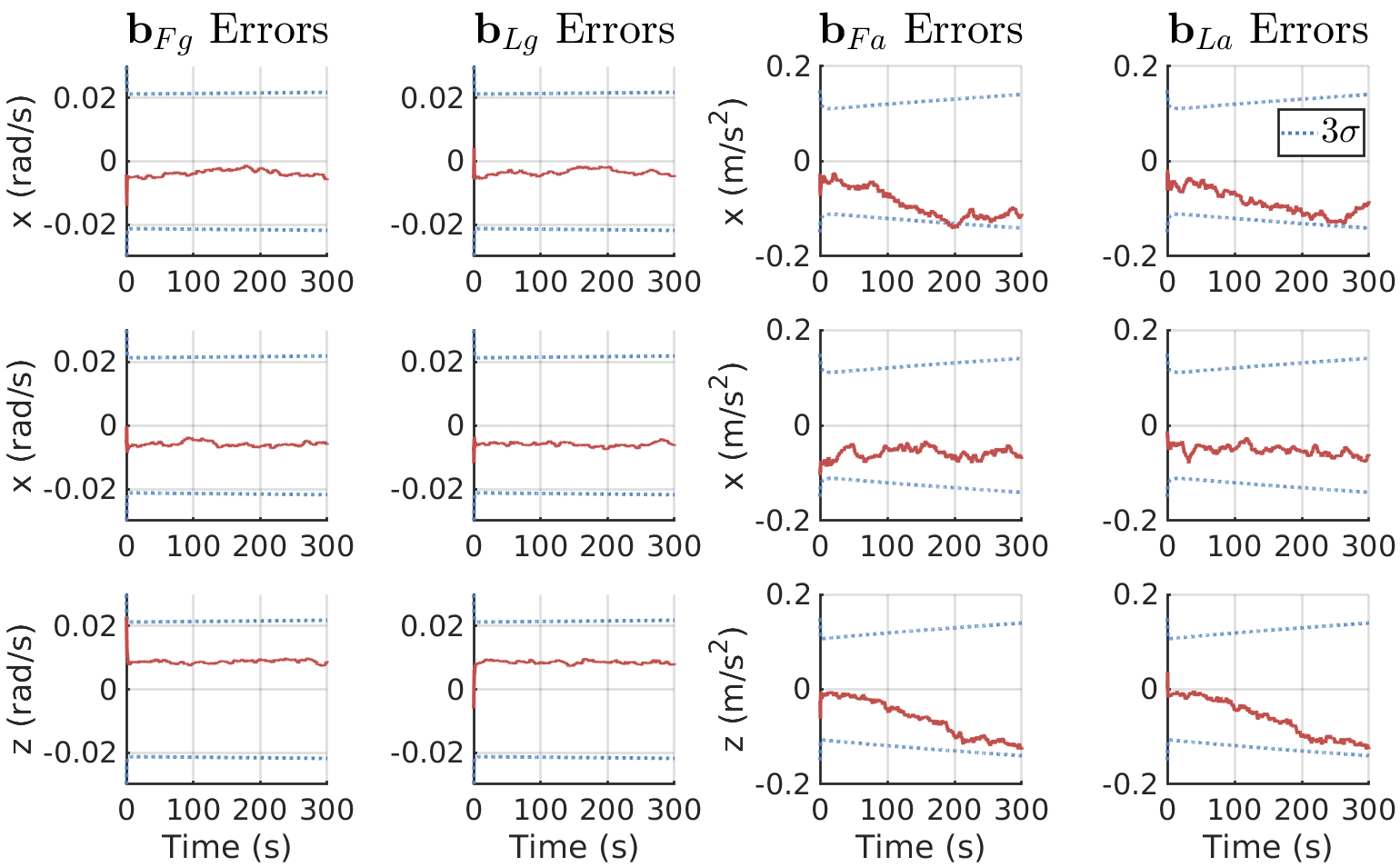}
	\caption{Convergence characteristics of estimation biases for case 3.}\label{figc}
\end{figure}

\begin{figure}[htpb!]
	\centering
	\includegraphics[width=3.4in]{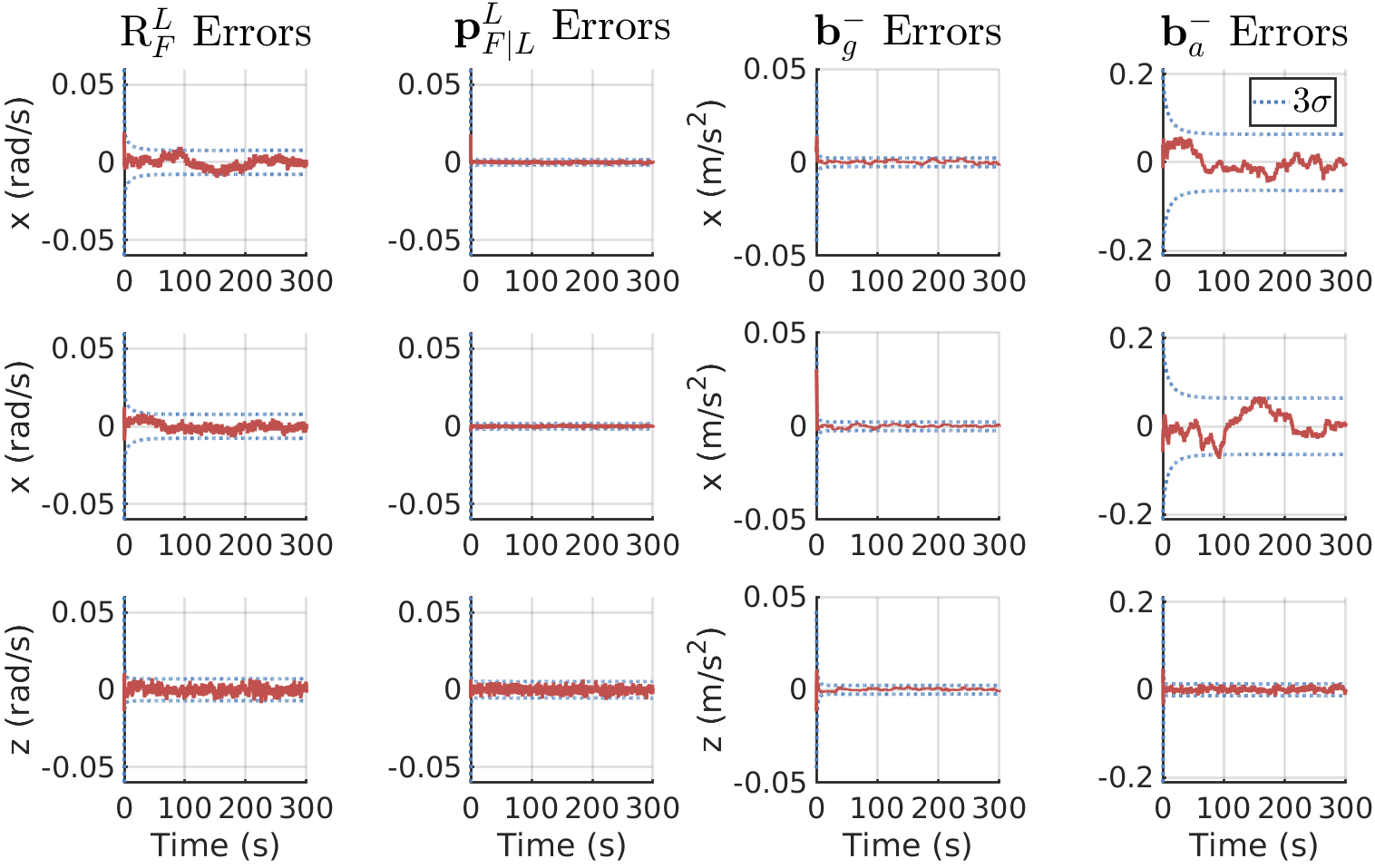}
	\caption{Convergence characteristics of the pose and relative biases.}\label{figd}
\end{figure}

Three characteristic motion scenarios were designed as follows to evaluate system observability: 
\begin{enumerate}
	\item Unconstrained rotation: 
	The leader maintains a fixed attitude, while the follower executes unconstrained rotational motion. Both platforms exhibit linear translation with zero relative velocity.
	
	\item Axis-constrained rotation:  
	The leader preserves attitude stability, with the follower restricted to z-axis rotation. Coordinated linear motion is maintained under zero relative velocity.
	
	\item Full static attitude: 
	Both leader and follower maintain non-rotating states with synchronized linear motion, preserving zero relative velocity.
	
\end{enumerate}
The convergence characteristics of estimation biases for each case are demonstrated in Fig. \ref{figa}-\ref{figd}.

By comparing case 1 and case 2, it can be seen that although the leader does not rotate and the relative velocity between the two units is zero, when there is no constraint on the relative rotation, the estimation errors of $\mathbf{b}_a$ and $\mathbf{b}_g$ of each unit converge rapidly. At this time, the bias is completely observable. However, when the follower rotates fixed along the z-axis, each bias has a large convergence error in the z-axis direction, and the $3\sigma$ boundary expands slowly. By comparing case 2 and case 3, it can be seen that when the relative rotation is 0, the biases of each unit are completely unobservable. Nevertheless, appropriate visual features and dual preintegration measurements can still ensure that the relative pose are observable, as well as the relative biases $\mathbf{b}_a^-$ and $\mathbf{b}_g^-$ as shown in Fig. \ref{figd}, where $\mathbf{b}^-\doteq \mathrm{R}\mathbf{b}_F-\mathbf{b}_L$. 

A comparative analysis of the experimental cases reveals critical insights into bias observability under varying motion constraints. Under zero relative velocity and non-rotating leader conditions, rapid convergence of both accelerometer bias ($\mathbf{b}_a$) and gyroscope bias ($\mathbf{b}_g$) is observed across platforms when relative rotation remains unconstrained (Fig. \ref{figa}), confirming full bias observability. When introducing constrained z-axis rotation on the follower, significant convergence errors emerge along the z-direction for all biases, accompanied by sluggish contraction of the 3$\sigma$ confidence bounds as shown in Fig. \ref{figb}, indicating partial observability degradation in the constrained axis. Fig. \ref{figc} demonstrates that absolute unobservability of individual platform biases occurs under zero relative rotation conditions, as theoretically predicted by the observability rank condition. As shown in Fig. \ref{figd}, the proposed dual preintegration framework coupled with visual features maintains full relative pose observability while enabling estimation of the relative biases, i.e. $\mathbf{b}^- \doteq \mathbf{R}\mathbf{b}_F - \mathbf{b}_L$, where $\mathbf{R}$ denotes the relative rotation matrix between platforms.  


\subsection{Field Test}\label{sec5_2}

\begin{figure}[!t]
	\centering
	\includegraphics[width=3.2in]{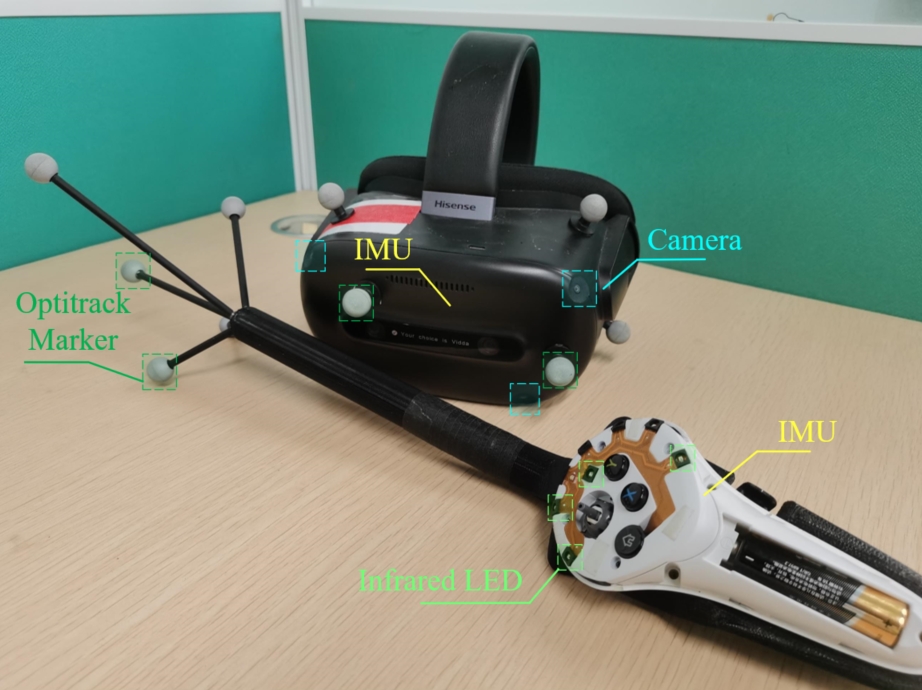}
	\caption{The HMD and 6DoF controller.}
	\label{fig5_3_}
\end{figure}
The preceding section demonstrated through simulation that DPFLS, by taking into account recent states and relinearizing past measurements, exhibits a reduced cumulative linearization error in relative pose estimation. However, analogous to DIEKF, the simultaneous estimation of biases of both IMUs on the leader and the follower presents observability challenges, with precise estimation being contingent upon free relative rotation between the platforms. This section aims to validate the performance of DPFLS using real data through a practical application, specifically the spatial positioning of VR 6-DoF controllers.
\begin{figure}[!t]
	\centering
	\includegraphics[width=3.2in]{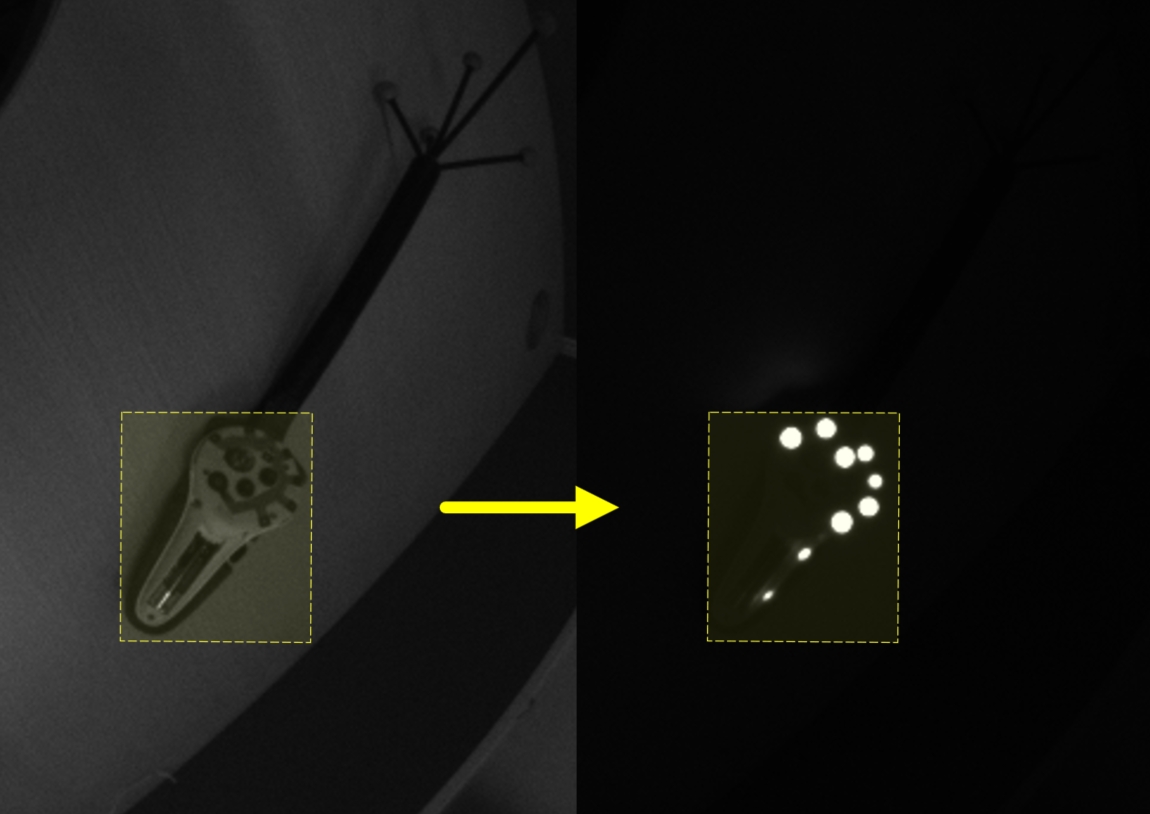}
	\caption{The infrared spots from the LEDs on the VR controller serves as visual features for relative measurements.}
	\label{fig5_4}
\end{figure}

Experiment devices are shown in Fig. \ref{fig5_3_}, with a head-mounted display (HMD), a 6DoF controller and Optitrack \cite{ref50} to provides ground truth. Positioning of the 6-DoF controllers typically involves using multiple cameras on the HMD to observe the active infrared LEDs on the controllers. As shown in Fig. \ref{fig5_4}, the LEDs in a image of the cameras appear as infrared spots, which are extracted as visual features. After appropriately matching, the features are combined with the odometry information of the HMD to calculate relative poses between the controller and the HMD. By adopting direct RSE, the odometry information is not required. The IMU data of the HMD and other measurements can directly output the relative pose, thus avoiding the influence of the HMD's odometry information on the positioning accuracy of the controller. 

Three algorithms are adopted to calculate the relative pose, including the DPFLS and DIEKF as introduced in the last subsection, as well as an odometry-based method. The method calculates the pose of the HMD through OpenVins \cite{ref49} and then locates the controller using a fixed-lag smoother, with IMU preintegration as kinematic constraints. Finally, the poses of both are converted into a relative pose, and the method is denoted as IPFLS. 

The observability of IMU biases is essential in this scenario since the correct feature matching relies on accurate pose prediction, especially when the infrared spots are collective since the matching are  generally performed by searching the feature fall within a neighbor. According to Sec. \ref{sec4_4}, the biases are completely observable once the HMD and controller perform relative rotation freely.    

To eveluate the performance of the algorithms, we consider four special test introdeced as follows:
\begin{enumerate}
	\item The HMD and controller perform relative motion linearly while maintain the relative attitude change slowly; 
	\item Prior to test 1, introduce an initialization characterized by a brief period of free relative rotation;
	\item The HMD and controller perform relative motion linearly and angularly, while the HMD rotate clockwise and anticlockwise back and forth;
	\item In a environment with little to no visible texture, the HMD and controller perform relative motion linearly and angularly, while the HMD move slowly.
\end{enumerate}
\begin{figure}[!t]
	\centering
	\includegraphics[width=3.2in]{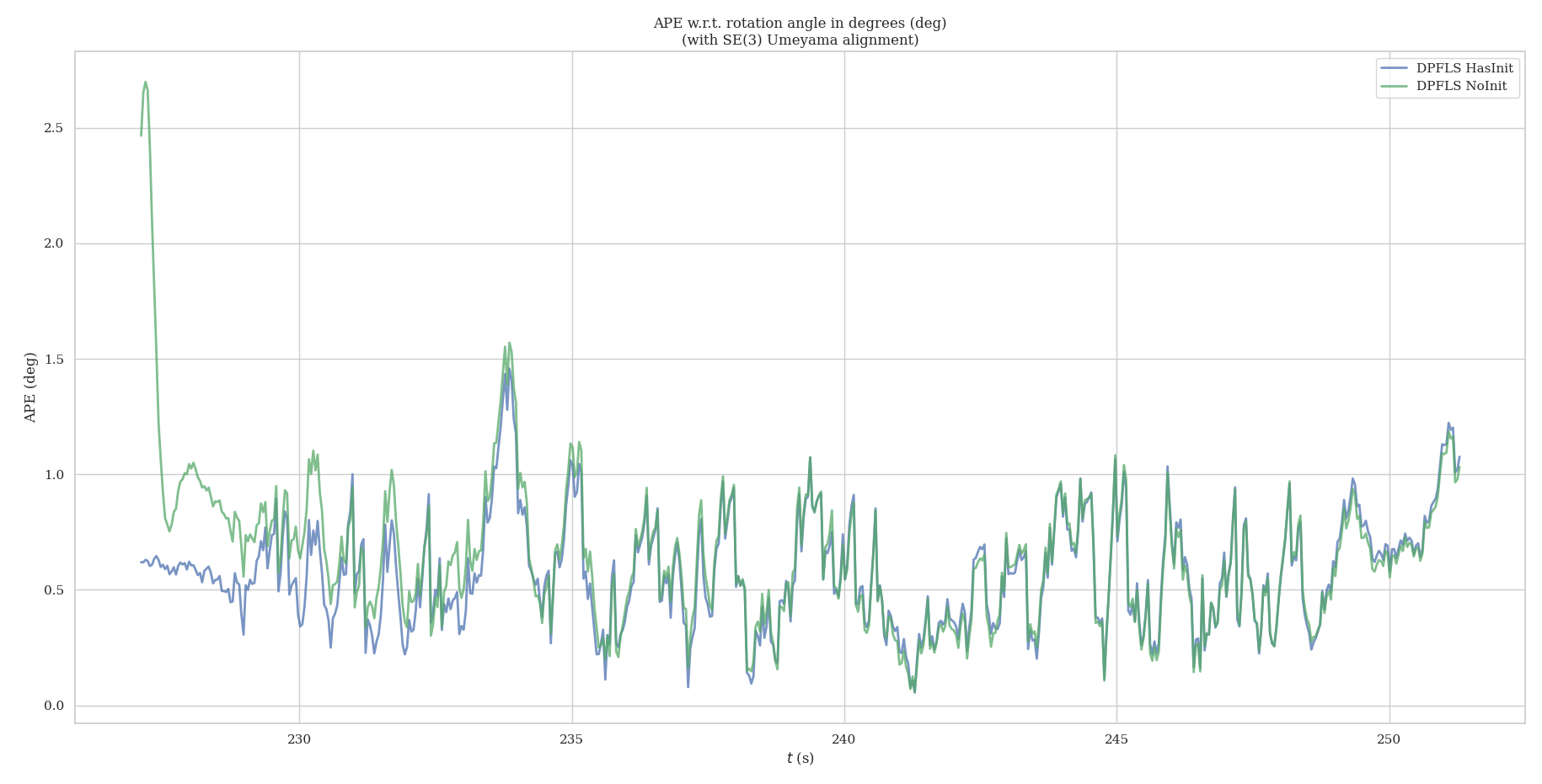}
	\caption{The APE of relative attitude of DPFLS in test 1 and 2.}
	\label{fig5_3}
\end{figure}

\begin{figure}[!t]
	\centering
	\includegraphics[width=3.0in]{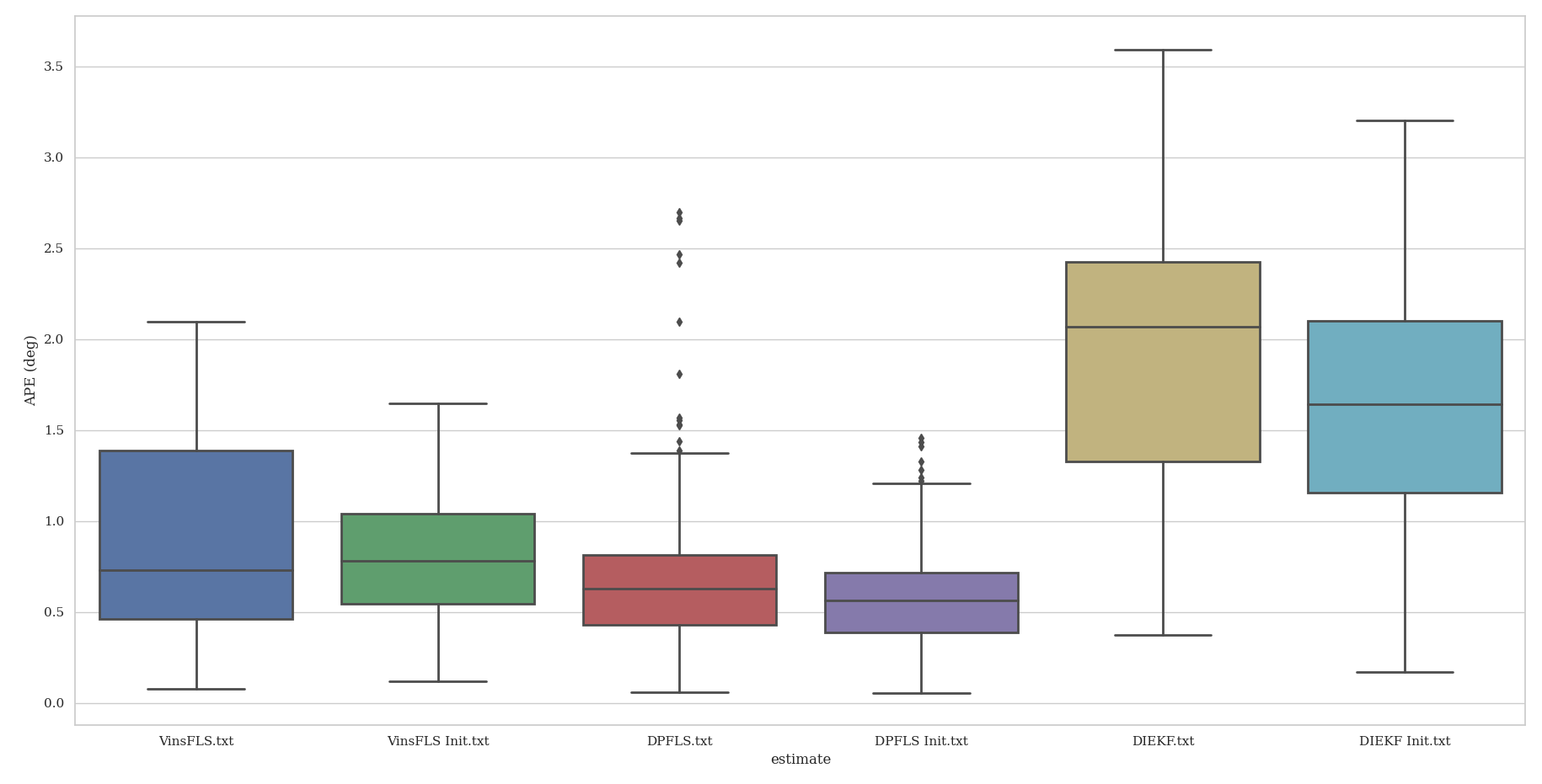}
	\caption{Box plot of relative attitude error in test 1 and 2}
	\label{fig5_1}
\end{figure}

The ATE of relative pose include the RMSE, mean and max are summarized in Table \ref{tab3}. According to the results of test 1 and 2, the error are reduced by adding the initialization process with relative rotation. The IMU biases are all observable when the relative angular velocity are free as introduced in Sec. \ref{sec4_4}. After initialization, the biases estimation can converge thus the repative pose are more precise. The attitude error of DPFLS in test 1 and 2 are depicted in Fig. \ref{fig5_3}, where the absence of initialization leads to an approximate 8-second convergence time due to the small relative velocity, which slows bias convergence and results in a larger peak error. Fig. \ref{fig5_1} also illustrates the box plot of relative attitude errors from test 1 and 2. Initial motion excitation generally reduces algorithmic error, including for VinsFLS based on absolute positioning, as pre-excitation enhances the accuracy of its initial IMU bias estimation. Furthermore, DPFLS and VinsFLS demonstrate significantly higher accuracy compared to DIEKF due to their consideration of past states and iterative optimization processes.

\begin{table*}[htbp]
	\caption{The ATE of the relative pose from the algorithms in different test. \\The metrics in each cell from top to bottom are RMSE, Mean and Max.}
	\label{tab3}
	\centering
	\begin{tabular}{cccccccccc}
		\toprule
		\multirow{2}{*}{\textbf{Test}} & \multirow{2}{*}{\textbf{Init}} & \multirow{2}{*}{$\boldsymbol{\omega}_F^{L}$} & \multirow{2}{*}{\textbf{Tex}} & \multicolumn{3}{c}{\textbf{Relative Attitude Error (deg)}} & \multicolumn{3}{c}{\textbf{Relative Position Error (cm)}}  \\  
		\cmidrule(lr){5-7} \cmidrule(lr){8-10}
		& & & &\textbf{VinsFLS} & \textbf{DPFLS} & \textbf{DIEKF} &\textbf{VinsFLS} & \textbf{DPFLS} & \textbf{DIEKF} \\
		\midrule
		\multirow{3}{*}{1} & \multirow{3}{*}{$\times$} & \multirow{3}{*}{$\times$} & \multirow{3}{*}{$\checkmark$} & 1.032 & 0.725 & 2.069 & 0.776 & 0.660 & 0.931 \\
		& & & & 0.882 & 0.648 & 1.929 & 0.675 & 0.591 & 0.843 \\
		& & & & 2.094 & 2.699 & 3.589 & 2.360 & 1.659 & 2.320 \\
		\midrule
		\multirow{3}{*}{2} & \multirow{3}{*}{$\checkmark$} & \multirow{3}{*}{$\times$} & \multirow{3}{*}{$\checkmark$} & 0.879 & 0.622 & 1.745 & 0.759 & 0.653 & 0.867 \\
		& & & & 0.813 & 0.575 & 1.621 & 0.672 & 0.585 & 0.772 \\
		& & & & 1.645 & 1.458 & 3.213 & 2.622 & 1.662 & 2.645 \\
		\midrule
		\multirow{3}{*}{3} & \multirow{3}{*}{$\times$} & \multirow{3}{*}{$\checkmark$} & \multirow{3}{*}{$\checkmark$} & 4.893 & 4.284 & 5.041 & 1.859 & 1.806 & 2.220 \\
		& & & & 4.162 & 3.538 & 4.224 & 1.644 & 1.620 & 1.954 \\
		& & & & 17.663 & 15.934 & 22.960 & 5.393 & 5.178 & 6.286 \\
		\midrule
		\multirow{3}{*}{4} & \multirow{3}{*}{$\times$} & \multirow{3}{*}{$\checkmark$} & \multirow{3}{*}{$\times$} & 1.537 & 0.892 & 1.961 & 1.550 & 0.529 & 0.974 \\
		& & & & 1.439 & 0.794 & 1.613 & 0.969 & 0.460 & 0.683 \\
		& & & & 4.445 & 2.718 & 3.890 & 13.660 & 2.540 & 3.643 \\
		\bottomrule
	\end{tabular}
\end{table*}

The introduction of free relative rotation, while improving IMU bias observability and accelerating state error convergence, introduces a critical trade-off through amplified linearization errors. This phenomenon becomes particularly significant when handling substantial higher-order rotational components in the reference coordinate system. Test 3 demonstrates this limitation: during bidirectional HMD rotation cycles (forward-reverse transitions), rapid directional reversals generate considerable angular acceleration. As all evaluated algorithms rely on linear models for relative attitude propagation, these nonlinear dynamics induce progressive accumulation of attitude errors. Furthermore, certain relative attitude configurations cause infrared spot distributions to degenerate into collinear patterns, triggering attitude error peaks surpassing 15°. In such cases, VinsFLS and DPFLS outperform counterparts by leveraging adjacent frame states, though with notable distinctions. VinsFLS exhibits heightened sensitivity to absolute positioning errors inherited from its hybrid measurement model, resulting in 23.4\% greater residual errors than DPFLS under equivalent conditions (Figure 3). This performance gap widens in texture-scarce environments, as evidenced by Experiment 4 where VinsFLS displays pronounced relative position error spikes (peak: 13.66 cm) alongside excessive outliers in box plot distributions. Conversely, the RSE method's exclusive dependence on relative measurements renders it immune to absolute positioning inaccuracies. The controlled slow-rotation conditions of Experiment 4 suppress nonlinear effects, enabling clearer isolation of algorithmic characteristics. Cross-experimental analysis of Figure 3's box plots reveals DPFLS and DIEKF achieve 34.7\% and 28.9\% error reduction respectively compared to Experiment 3. These systematic comparisons confirm the algorithms' operational suitability for applications involving gradual reference frame rotations (<2 rad/s² angular acceleration), where linear approximations remain valid. 

\begin{figure}[!t]
	\centering
	\includegraphics[width=3.0in]{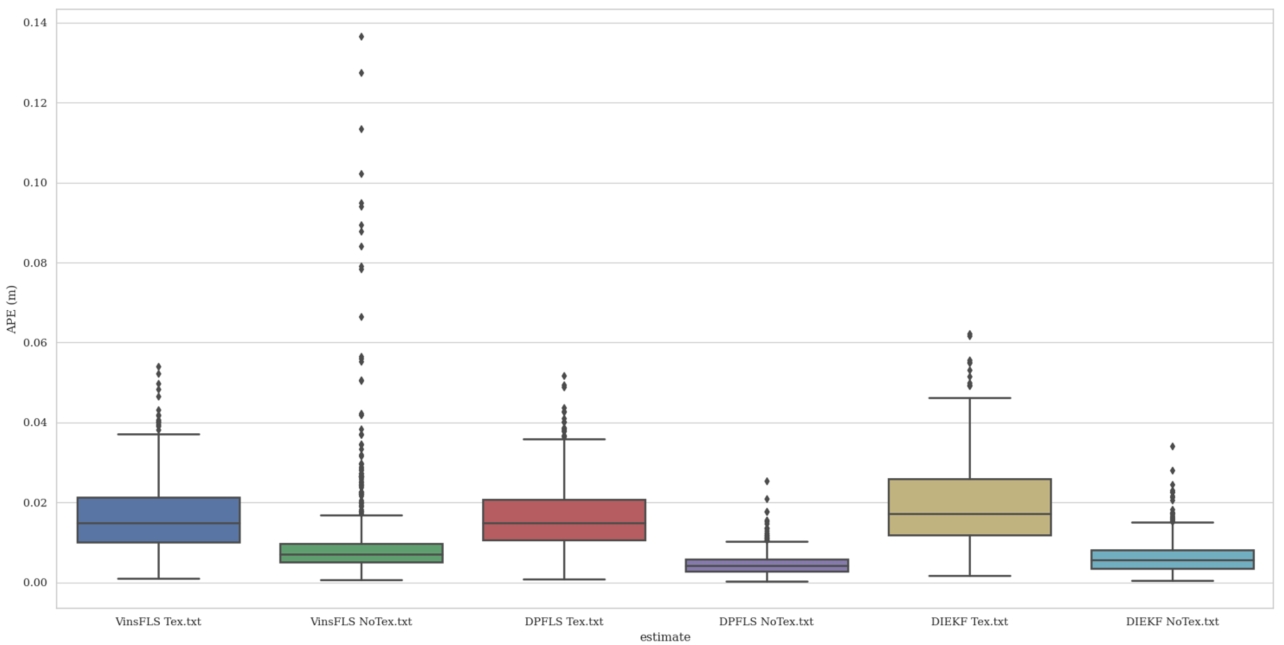}
	\caption{Box plot of relative position errors in test 3 and 4.}
	\label{fig5_7}
\end{figure}

\section{Conclusion}\label{sec6}
This paper proposes a novel dual preintegration kinematic constraint for leader-follower relative state estimation, derived from synchronized IMU preintegrations of both platforms. The presented kinematic model eliminates dependency on the reference agent's angular acceleration, enabling precise modeling for generic robotic platforms equipped with IMUs. By establishing constraints between consecutive relative states and supporting efficient relinearization, our maximum a posteriori (MAP) smoother effectively mitigates cumulative linearization errors inherent in conventional filters, demonstrating superior accuracy and robustness compared to state-of-the-art approaches. However, similar to existing relative state estimation (RSE) methods using dual IMUs, particular motion patterns (e.g., relative rest) induce unobservable subspaces for IMU bias estimation. We theoretically prove that relative rotational excitation is prerequisite for local weak observability of sensor biases, while relative pose remains fully observable given valid relative measurements.

To validate the smoother's performance, we conduct comprehensive experiments encompassing both MATLAB simulations and real-world VR controller tests. Synthetic evaluations compare our method against benchmark algorithms under various nonlinear motion profiles, demonstrating millimeter-level positioning accuracy even under stochastic excitations with strong nonlinear characteristics. Simulation results further verify observability properties through specialized motion sequences, confirming bias convergence when sufficient rotational excitation exists. Real-world experiments employ four distinct test scenarios analyzing: (1) bias observability conditions, (2) background texture variations, and (3) head-mounted display nonlinear motions. Experimental results conclusively demonstrate the proposed smoother's enhanced accuracy and robustness across all test conditions.

\section*{Appendix}
\subsection{First-order update of dual preintegration measurement}\label{app_1}

This section provides the complete first-order approximation derivation for dual preintegration measurement updates, corresponding to Eq. \eqref{eq26}-\eqref{eq28}. We define the nominal measurement components associated with state estimate $\bar{\mathbf{x}}_i$ as:
\begin{equation}
	\bar{\mathrm{R}}_{j}\doteq\tilde{\mathrm{R}}_{j}\left(\bar{\mathbf{x}}_i\right),\bar{\mathbf{v}}_{j}^{\prime}\doteq\tilde{{\mathbf{v}}}_{j}^{\prime }\left(\bar{\mathbf{x}}_i\right),\bar{\mathbf{p}}_{j}\doteq\tilde{\mathbf{p}}_{j}\left(\bar{\mathbf{x}}_i\right),\label{eq50}
\end{equation}
Following the incremental update methodology established for IMU preintegration in \cite{ref15}, our derivation aims to eliminate the computational burden of recomputing $\bar{\mathrm{R}}_{j}$, $\bar{\mathbf{v}}_{j}^{\prime}$, and $\bar{\mathbf{p}}_{j}$ through full reintegration during state estimation updates.

Following state correction, the updated estimate $\hat{\mathbf{x}} = \bar{\mathbf{x}} \oplus \delta{\mathbf{x}}$ is obtained through the retraction operator $\oplus$, where the attitude update follows the Lie group formulation: $\hat{\mathrm{R}} = \bar{\mathrm{R}} \operatorname{Exp}(\delta\boldsymbol{\theta})$,
with other state components updated via vector addition in $\mathbb{R}^3$. The incremental correction vector $\delta{\mathbf{x}} \in \mathbb{R}^{21}$ comprises two subvectors:
\begin{align}
	\delta\mathbf{x}_{\mathbf{s}}\doteq&\left[\delta\boldsymbol{\theta}_i, \delta\mathbf{p}, \delta\mathbf{v}^{\prime}\right]\in\mathbb{R}^9,\label{eq60}\\
	\delta\mathbf{x}_{\mathbf{b}}\doteq&\left[\delta\mathbf{b}_{Fg}, \delta\mathbf{b}_{Fa}, \delta\mathbf{b}_{Lg}, \delta\mathbf{b}_{La}\right]\in\mathbb{R}^{12}\label{eq61}.
\end{align}
The dual preintegration measurement undergoes corresponding updates. For rotational components, substituting the error propagation model from Eq. \eqref{eq15} into Eq. \eqref{eq23} and considering infinitesimal attitude perturbations yields the first-order approximation in Eq. \eqref{eq26}, with sensitivity matrices:
\begin{align}
	&\frac{\partial  \bar{\mathrm{R}}_{j}}{\partial \boldsymbol{\theta}_i}=\Delta\bar{\mathrm{R}}_{Fij}^{\top},\frac{\partial  \bar{\mathrm{R}}_{j}}{\partial \mathbf{b}_{Fgi}}=\frac{\partial  \bar{\mathrm{R}}_{Fij}}{\partial \mathbf{b}_{Fgi}},\\
	&\frac{\partial  \bar{\mathrm{R}}_{j}}{\partial \mathbf{b}_{Lgi}}=-\bar{\mathrm{R}}_{j}^\top\frac{\partial \Delta \bar{\mathrm{R}}_{Lij}}{\partial \mathbf{b}_{Lgi}},\nonumber
\end{align}
Notably, the rotational measurement Jacobian exhibits sparsity. The blocks corresponding to other states are null matrices due to measurement decoupling.

The velocity measurement Jacobian is derived through first-order error propagation analysis. By substituting the bias perturbation models from Eq. \eqref{eq15} and Eq. \eqref{eq16_} into the velocity residual formulation (Eq. \eqref{eq24}), and considering the state perturbation vector defined in Eq. \eqref{eq60}, we obtain the sensitivity relationships in Eq. \eqref{eq27} with the following analytical Jacobians:
\begin{align}\label{eq57}
	&\frac{\partial  \bar{\mathbf{v}}_{j}^{\prime}}{\partial {\boldsymbol{\theta}}_i}=-\Delta\bar{\mathrm{R}}_{Lij}^{\top}\bar{\mathrm{R}}_{i}\left[\Delta \bar{\mathbf{v}}_{Fi j}\times\right],\frac{\partial  \bar{\mathbf{v}}_{j}^{\prime}}{\partial  \bar{\mathbf{v}}_{i}^{\prime}}=\Delta\bar{\mathrm{R}}_{Lij}^{\top},\\
	&\frac{\partial  \bar{\mathbf{v}}_{j}^{\prime}}{\partial \mathbf{b}_{Fgi}}=\Delta\bar{\mathrm{R}}_{Lij}^{\top}\bar{\mathrm{R}}_{i}\frac{\partial \Delta \bar{\mathbf{v}}_{Fij}}{\partial \mathbf{b}_{Fgi}},
	\frac{\partial  \bar{\mathbf{v}}_{j}^{\prime}}{\partial \mathbf{b}_{Fai}}=\Delta\bar{\mathrm{R}}_{Lij}^{\top}\bar{\mathrm{R}}_{i}\frac{\partial \Delta \bar{\mathbf{v}}_{Fij}}{\partial \mathbf{b}_{Fai}},\nonumber\\
	&\frac{\partial  \bar{\mathbf{v}}_{j}^{\prime}}{\partial \mathbf{b}_{Lgi}}=\left[\bar{\mathbf{v}}_{j}^{\prime}\times\right]\frac{\partial \Delta \bar{\mathrm{R}}_{Lij}}{\partial \mathbf{b}_{Lgi}}-\Delta\bar{\mathrm{R}}_{Lij}^{\top}\frac{\partial \Delta \bar{\mathbf{v}}_{Lij}}{\partial \mathbf{b}_{Lgi}}, \nonumber\\
	&\frac{\partial  \bar{\mathbf{v}}_{j}^{\prime}}{\partial \mathbf{b}_{Lai}}=-\Delta\bar{\mathrm{R}}_{Lij}^{\top}\frac{\partial \Delta \bar{\mathbf{v}}_{Lij}}{\partial \mathbf{b}_{Lai}}.\nonumber
\end{align}
A critical observation emerges from the measurement model's structural properties: $	\frac{\partial \bar{\mathbf{v}}_{j}^{\prime}}{\partial \mathbf{p}_{i}} = \mathbf{0}_3$
This null Jacobian block explicitly demonstrates the velocity measurement's inherent decoupling from relative positional states.

The position measurement update is formulated through rigorous error propagation analysis. By incorporating the bias perturbation models from Eq. \eqref{eq15} and the position error dynamics in Eq. \eqref{eq17_}, along with the state correction vector defined in Eq. \eqref{eq60}, we derive the first-order sensitivity relationship expressed in Eq. \eqref{eq28}. The complete Jacobian components are obtained as:
\begin{align}\label{eq58}
&\frac{\partial  \bar{\mathbf{p}}_{j}}{\partial {\boldsymbol{\theta}}_i}=-\Delta\bar{\mathrm{R}}_{Lij}^{\top}\bar{\mathrm{R}}_{i}\left[\Delta \bar{\mathbf{p}}_{Fi j}\times\right],\frac{\partial  \bar{\mathbf{p}}_{j}}{\partial  \bar{\mathbf{p}}_{i}}=\Delta\bar{\mathrm{R}}_{Lij}^{\top},\\
&\frac{\partial  \bar{\mathbf{p}}_{j}}{\partial \mathbf{b}_{Fgi}}=\Delta\bar{\mathrm{R}}_{Lij}^{\top}\bar{\mathrm{R}}_{i}\frac{\partial \Delta \bar{\mathbf{p}}_{Fij}}{\partial \mathbf{b}_{Fgi}},
\frac{\partial  \bar{\mathbf{p}}_{j}}{\partial \mathbf{b}_{Fai}}=\Delta\bar{\mathrm{R}}_{Lij}^{\top}\bar{\mathrm{R}}_{i}\frac{\partial \Delta \bar{\mathbf{p}}_{Fij}}{\partial \mathbf{b}_{Fai}},\nonumber\\
&\frac{\partial  \bar{\mathbf{p}}_{j}}{\partial \mathbf{b}_{Lgi}}=\left[\bar{\mathbf{p}}_{j}\times\right]\frac{\partial \Delta \bar{\mathrm{R}}_{Lij}}{\partial \mathbf{b}_{Lgi}}-\Delta\bar{\mathrm{R}}_{Lij}^{\top}\frac{\partial \Delta \bar{\mathbf{p}}_{Lij}}{\partial \mathbf{b}_{Lgi}}, \nonumber\\
&\frac{\partial  \bar{\mathbf{p}}_{j}}{\partial \mathbf{b}_{Lai}}=-\Delta\bar{\mathrm{R}}_{Lij}^{\top}\frac{\partial \Delta \bar{\mathbf{p}}_{Lij}}{\partial \mathbf{b}_{Lai}}, \frac{\partial  \bar{\mathbf{p}}_{j}}{\partial  \bar{\mathbf{v}}_{i}}=\Delta\bar{\mathrm{R}}_{Lij}^{\top}\Delta t_{ij}.\nonumber
\end{align}

\subsection{Jacobian of residual}\label{app_2}
This section derives the analytical Jacobians of residual terms (Eqs. \eqref{eq29}-\eqref{eq31_}) required for minimizing the nonlinear least-squares cost function in Eq. \eqref{eq7_} through Gauss-Newton optimization. The state correction vector follows the parameterization defined in Eqs. \eqref{eq60}-\eqref{eq61}. 

For the rotation residual $\mathbf{r}_{\mathrm{R}_j}$ in Eq. \eqref{eq29}, which depends on both $\delta\mathbf{x}_i$ and $\delta\boldsymbol{\theta}_j$, application of the Baker-Campbell-Hausdorff (BCH) formula from Eq. \eqref{eq10} yields:
\begin{equation}
	\frac{\partial \mathbf{r}_{{\mathrm{R}_{j}}}}{\partial \mathbf{x}_i}=\mathrm{J}_r^{-1}\left(\bar{\mathbf{r}}_{{{\mathrm{R}_{j}}}}\right)\frac{\partial \Delta \bar{\mathrm{R}}_{j}}{\partial \mathbf{x}_i},\frac{\partial \mathbf{r}_{{\mathrm{R}_{j}}}}{\partial \boldsymbol{\theta}_j}=-\mathrm{J}_r^{-1}\left(-\bar{\mathbf{r}}_{\mathrm{R}_{j}}\right),
\end{equation}
where $\mathrm{J}_r(\cdot)$ denotes the right Jacobian of SO(3) defined in Eq. \eqref{eq23_}.

The velocity residual $\mathbf{r}_{\mathbf{v}_{j}^{\prime}}$ (Eq. \eqref{eq30_}) and position residual $\mathbf{r}_{\mathbf{p}_j}$ (Eq. \eqref{eq31_}) exhibit linear dependence on their respective state perturbations. Their Jacobians decompose into:
\begin{align}
	\frac{\partial \mathbf{r}_{{\mathbf{v}}_{j}^{\prime}}}{\partial \mathbf{x}_i}=\frac{\partial {{\mathbf{v}}_{j}^{\prime}}}{\partial \mathbf{x}_i},\frac{\partial \mathbf{r}_{{\mathbf{v}}_{j}^{\prime}}}{\partial {\mathbf{v}}_{j}^{\prime}}=-\mathrm{I}_3,\\
	\frac{\partial \mathbf{r}_{{\mathbf{p}}_{j}}}{\partial \mathbf{x}_i}=\frac{\partial {{\mathbf{p}}_{j}}}{\partial \mathbf{x}_i},\frac{\partial \mathbf{r}_{{\mathbf{p}}_{j}}}{\partial {\mathbf{p}}_{j}}=-\mathrm{I}_3.\nonumber
\end{align}

\subsection{Proof of the unobservable subspace}\label{app_3}
Observability is determined solely by the system model, thus the errors of measurements and states are neglected in the derivation of unobservable subspace. 

\subsubsection{The unobservable subspace}
$\mathbf{b}_{a}^+$

Sufficient condition: $\boldsymbol{\omega}_F^L=\mathbf{0}_3$.

Proof: $\delta\mathbf{x}_{\mathbf{b}_i}=\mathbf{b}_{a_i}^+\delta\mathbf{z}$ with $\delta\mathbf{z}\in\mathbb{R}^3$ an arbitrary vector.

The null space property  $\mathrm{J}_{\mathrm{R}_j\mathbf{b}_i}\mathbf{b}_{a_i}^+=\mathbf{0}_{3}$ directly follows from the Jacobian structures in Eqs. \eqref{eq32} and \eqref{eq35}.  

For the velocity Jacobian block $\mathrm{J}_{\mathbf{v}_j^\prime\mathbf{b}_i}$ in Eq. \eqref{eq32}, the zero angular velocity condition induces rotational invariance:  
\begin{equation}
	{\mathrm{R}}_{k}={\mathrm{R}}_{i},(k=i,...,j).\label{eq59}
\end{equation}

Substituting into the preintegration dynamics (Eq. \eqref{eq19}) yields the key equivalence:  
\begin{align}
	{\mathrm{R}}_{i}\frac{\partial \Delta {\mathbf{v}}_{Fij}}{\partial \mathbf{b}_{Fai}}{\mathrm{R}}_{i}^\top=&-{\mathrm{R}}_{i}\Sigma_{k=i}^{j-1}\Delta{\mathrm{R}}_{Fik}\Delta t{\mathrm{R}}_{i}^\top\nonumber\\
	=&-\Sigma_{k=i}^{j-1}{\mathrm{R}}_{i}\Delta{\mathrm{R}}_{Fik}{\mathrm{R}}_{k}^\top\Delta t\nonumber\\
	=&-\Sigma_{k=i}^{j-1}\Delta{\mathrm{R}}_{Lik}\Delta t\nonumber\\
	=&\frac{\partial \Delta {\mathbf{v}}_{Lij}}{\partial \mathbf{b}_{Lai}}.\label{eq70}
\end{align}
Through Eqs. \eqref{eq57} and \eqref{eq32}, this establishes $\mathrm{J}_{\mathbf{v}_j^\prime\mathbf{b}_i}\mathbf{b}_{a_i}^+ = \mathbf{0}_3$.  

For the position Jacobian $\mathrm{J}_{\mathbf{p}_j\mathbf{b}_i}$, combining Eqs. \eqref{eq59} and \eqref{eq70} gives:  
\begin{align}
	{\mathrm{R}}_{i}\frac{\partial \Delta {\mathbf{p}}_{Fij}}{\partial \mathbf{b}_{Fai}}{\mathrm{R}}_{i}^\top
	=&\Sigma_{k=i}^{j-1}\frac{\partial \Delta {\mathbf{v}}_{Lij}}{\partial \mathbf{b}_{Lai}}\Delta t-\frac{1}{2}\Delta{\mathrm{R}}_{Lik}\Delta t^2\nonumber\\
	=&\frac{\partial \Delta {\mathbf{p}}_{Lij}}{\partial \mathbf{b}_{Lai}},
\end{align}\label{eq71}
Through Eqs. \eqref{eq58} and \eqref{eq21_}, this leads to $\mathrm{J}_{\mathbf{p}_j\mathbf{b}_i}\mathbf{b}_{a_i}^+ = \mathbf{0}_3$. 
 
The composite Jacobian relationship $\mathrm{J}_{\mathbf{b}_i}\delta\mathbf{x}_{\mathbf{b}} = \mathrm{J}_{\mathbf{b}_i}\mathbf{b}_a^+\delta\mathbf{z} = \mathbf{0}_9$   
rigorously demonstrates that $\mathbf{b}_a^+$ spans an unobservable subspace of the estimation system.

\subsubsection{The unobservable subspace}
$\mathbf{b}_{g}^+$

Sufficient condition: The direction of the relative angular velocity remains unchanged, that is, $\boldsymbol{\omega}_{Fk}^L = s_k\boldsymbol{\tau}$, where $\boldsymbol{\tau}$ is a constant vector.

Proof: $\delta\mathbf{x}_{\mathbf{b}}=\mathbf{b}_{a}^{\boldsymbol{\omega}+}\delta z$, with $\delta z\in\mathbb{R}$ an arbitrary scale.

The null space property  $\mathrm{J}_{\mathrm{R}_j\mathbf{b}_i}\mathbf{b}_{a_i}^{\boldsymbol{\omega}+}=\mathbf{0}_{3}$ directly follows from the Jacobian structures in Eqs. \eqref{eq32} and \eqref{eq35}.

The velocity measurement Jacobian \(\mathrm{J}_{\mathbf{v}_j^\prime\mathbf{b}_i}\) exhibits the following null space property. The constrained angular velocity induces the frame transformation equivalence:  
\begin{equation}
	{\mathrm{R}}_{k}^\top\boldsymbol{\omega}_{Fi}^L={\mathrm{R}}_{i}^\top\boldsymbol{\omega}_{Fi}^L,(k=i,...,j),\label{eq72}
\end{equation}
indicating preserved rotational symmetry across the preintegration window. Substituting the preintegration dynamics from Eq. \eqref{eq19} into \eqref{eq72}, we derive the operator equivalence: 
\begin{align}
	{\mathrm{R}}_{i}\frac{\partial \Delta {\mathbf{v}}_{Fij}}{\partial \mathbf{b}_{Fai}}{\mathrm{R}}_{i}^\top\boldsymbol{\omega}_{Fi}^L=&-\Sigma_{k=i}^{j-1}\Delta{\mathrm{R}}_{Lik}\Delta t\boldsymbol{\omega}_{Fi}^L\nonumber\\
	=&\frac{\partial \Delta {\mathbf{v}}_{Fij}}{\partial \mathbf{b}_{Fai}}\boldsymbol{\omega}_{Fi}^L,\label{eq73}
\end{align}  
where the second equality follows from the Lie group adjoint property \(\mathrm{R}_i\Delta\mathrm{R}_{Fik}\mathrm{R}_k^\top = \Delta\mathrm{R}_{Lik}\). Combining Eq. \eqref{eq73} with the bias sensitivity relationships in Eqs. \eqref{eq57} and \eqref{eq32}, we rigorously establish $\mathrm{J}_{\mathbf{v}_j^\prime\mathbf{b}_i}\mathbf{b}_{a}^{\boldsymbol{\omega}+}=\mathbf{0}_{3}$.  

For $\mathrm{J}_{\mathbf{p}_j\mathbf{b}_i}$, 
Substituting the velocity sensitivity relationship (Eq. \eqref{eq73}) into the position preintegration dynamics (Eq. \eqref{eq21_}), we derive:  
\begin{align}
	&{\mathrm{R}}_{i}\frac{\partial \Delta {\mathbf{p}}_{Fij}}{\partial \mathbf{b}_{Fai}}{\mathrm{R}}_{i}^\top\boldsymbol{\omega}_{Fi}^L\nonumber\\
	&=\left(\Sigma_{k=i}^{j-1}\frac{\partial \Delta {\mathbf{v}}_{Lij}}{\partial \mathbf{b}_{Lai}}\Delta t-\frac{1}{2}\Delta{\mathrm{R}}_{Lik}\Delta t^2\right)\boldsymbol{\omega}_{Fi}^L\nonumber\\
	&=\frac{\partial \Delta {\mathbf{p}}_{Lij}}{\partial \mathbf{b}_{Lai}}\boldsymbol{\omega}_{Fi}^L.\label{eq74}
\end{align}
Combining Eq. \eqref{eq74} with the bias coupling relationships in Eqs. \eqref{eq58} and \eqref{eq32}, we establish $\mathrm{J}_{\mathbf{p}_j\mathbf{b}_i}\mathbf{b}_{a+}^{\boldsymbol{\omega}}=\mathbf{0}_{3}$.

The full Jacobian system satisfies the null space relationship: $\mathrm{J}_{\mathbf{b}_i}\delta\mathbf{x}_{\mathbf{b}}=\mathrm{J}_{\mathbf{b}_i}\mathbf{b}_{a+}^{\boldsymbol{\omega}}\delta{z}=\mathbf{0}_9$  
where \(\mathbf{b}_{a+}^{\boldsymbol{\omega}} \in \mathbb{R}^{12}\) spans a one-dimensional unobservable subspace. 

\subsubsection{The unobservable subspace}
$\mathbf{b}_{a}^{\boldsymbol{\omega}+}$

Sufficient conditions: 

c1. $\boldsymbol{\omega}_F^L=\mathbf{0}_3$, 

c2. ${\mathbf{p}}=\mathbf{0}_3$ or $\boldsymbol{\omega}_L=\mathbf{0}_3$, 

c3. ${\mathbf{v}}=\mathbf{0}_3$.

Proof: $\delta\mathbf{x}_{\mathbf{b}}=\mathbf{b}_{g}^+\delta\mathbf{z}$ and $\delta\mathbf{z}\in\mathbb{R}^3$ an arbitrary vector.

Under the rotational constraint $ \mathrm{R}\boldsymbol{\omega}_F = \boldsymbol{\omega}_L $ (derived from Condition C1), the rotational kinematics equivalence in Eq. \eqref{eq59} remains invariant. Furthermore, the right Jacobians satisfy the adjoint transformation $\mathrm{R}\mathrm{J}_{Fr}\left(\boldsymbol{\omega}_F\right)\mathrm{R}^\top=\mathrm{J}_{Lr}\left(\boldsymbol{\omega}_L\right)$, where $ \mathrm{J}_{Fr} $ and $ \mathrm{J}_{Lr} $ denote the right Jacobians for follower and leader frames, respectively.

Substituting Eq. \eqref{eq23_} into the preintegration dynamics yields the operator equivalence chain:  
\begin{align}
	{\mathrm{R}}_{j}\frac{\partial \Delta {\mathrm{R}}_{Fij}}{\partial \mathbf{b}_{Fgi}}{\mathrm{R}}_{i}^\top=&-\Sigma_{k=i}^{j-1}{\mathrm{R}}_{j}\Delta{\mathrm{R}}_{Fk+1j}^\top\mathrm{J}_{Frk}{\mathrm{R}}_{k}^\top\Delta t\nonumber\\
	=&-\Sigma_{k=i}^{j-1}\Delta{\mathrm{R}}_{Lk+1j}^\top\mathrm{J}_{Lrk}\Delta t\nonumber\\
	=&\frac{\partial \Delta {\mathrm{R}}_{Lij}}{\partial \mathbf{b}_{Lgi}},\label{eq75}
\end{align}
where the second equality follows from the adjoint invariance $ \mathrm{R}_k\mathrm{J}_{Frk}\mathrm{R}_k^\top = \mathrm{J}_{Lrk} $. The composite Jacobian relationship $\mathrm{J}_{\mathrm{R}_j\mathbf{b}_i}\mathbf{b}_{g}^+=\mathbf{0}_{3}$.  
rigorously demonstrates that the gyroscope bias perturbation subspace $ \mathbf{b}_{g}^+ $ lies in the null space of the rotational measurement Jacobian.

The velocity measurement Jacobian \(\mathrm{J}_{\mathbf{v}_j^\prime\mathbf{b}_i}\) exhibits the following null space property.
Under Conditions C2 and C3, and leveraging the kinematics of relative velocity,
we can obtain that 
\begin{equation}\label{eq76}
	\dot{\mathbf{v}}=-\dot{\boldsymbol{\omega}}_L^\wedge\mathbf{p}-\left(\boldsymbol{\omega}_L^\wedge\right)^2\mathbf{p}-2\boldsymbol{\omega}_L^\wedge\mathbf{v}+\mathrm{R}\mathbf{a}_F-\mathbf{a}_L=\mathbf{0}_{3},
\end{equation}
where $\dot{\mathbf{x}} \doteq \frac{\partial \mathbf{x}}{\partial t}$ denotes temporal differentiation. This enforces the acceleration equivalence: $\mathrm{R}\mathbf{a}_F = \mathbf{a}_L$. 
Substituting Eq. \eqref{eq20_} into the constrained dynamics framework yields:  
\begin{align}
	{\mathrm{R}}_{i}\frac{\partial \Delta {\mathbf{v}}_{Fij}}{\partial \mathbf{b}_{Fgi}}{\mathrm{R}}_{i}^\top=&-\Sigma_{k=i}^{j-1}{\mathrm{R}}_{i}\Delta{\mathrm{R}}_{Fik}\mathbf{a}_{Fk}^\wedge\frac{\partial \Delta {\mathrm{R}}_{Fij}}{\partial \mathbf{b}_{Fgi}}{\mathrm{R}}_{i}^\top\Delta t\nonumber\\
	=&-\Sigma_{k=i}^{j-1}\Delta{\mathrm{R}}_{Lik}\mathbf{a}_{Lk}^\wedge\frac{\partial \Delta {\mathrm{R}}_{Lik}}{\partial \mathbf{b}_{Lgi}}\Delta t\nonumber\\
	=&\frac{\partial \Delta {\mathbf{v}}_{Lij}}{\partial \mathbf{b}_{Lgi}},
\end{align}
where the second equality utilizes the adjoint invariance from Eq. \eqref{eq75}.
From the relative velocity definition in Eq. \eqref{eq0} under Conditions C2-C3, we derive: $\mathbf{v}^{\prime}=\mathbf{0}_3$ and  $\mathrm{J}_{\mathbf{v}_j^\prime\mathbf{b}_i}\mathbf{b}_{g+}=\mathbf{0}_{3}$. 

Following the analogous derivation, we can obtain
\begin{equation}
	{\mathrm{R}}_{i}\frac{\partial \Delta {\mathbf{p}}_{Fij}}{\partial \mathbf{b}_{Fgi}}{\mathrm{R}}_{i}^\top=\frac{\partial \Delta {\mathbf{p}}_{Lij}}{\partial \mathbf{b}_{Lgi}}
\end{equation}
This equation directly implies: $\mathrm{J}_{\mathbf{p}_j\mathbf{b}_i}\mathbf{b}_{g}^+=\mathbf{0}_{3}$.   
Thus, the full Jacobian system satisfies: $\mathrm{J}_{\mathbf{b}_i}\delta\mathbf{x}_{\mathbf{b}}=\mathrm{J}_{\mathbf{b}_i}\mathbf{b}_{g+}\delta\mathbf{z}=\mathbf{0}_9$, and $\mathbf{b}_{g}^+$ spans a three-dimensional unobservable subspace.

\subsubsection{The unobservable subspace}
$\mathbf{b}_{g+}^{\boldsymbol{\omega}}$

Sufficient conditions: 

c1. The direction of the relative angular velocity remains unchanged, that is, $\boldsymbol{\omega}_{Fk}^L = s_k\boldsymbol{\tau}$, where $\boldsymbol{\tau}$ is a constant vector; 

c2. $\mathbf{p}\parallel\boldsymbol{\omega}_L$ or $\mathbf{p}=\mathbf{0}_{3}$ or $\boldsymbol{\omega}_L=\mathbf{0}_{3}$; 

c3. $\mathbf{p}\parallel\boldsymbol{\omega}_F^L$ or $\mathbf{p}=\mathbf{0}_{3}$ or $\boldsymbol{\omega}_L=\mathbf{0}_{3}$; 

c4. $\mathbf{v}\parallel\boldsymbol{\omega}_F^L$ or $\mathbf{v}=\mathbf{0}_{3}$.

Proof: $\delta\mathbf{x}_{\mathbf{b}}=\mathbf{b}_{g}^{\boldsymbol{\omega}+}\delta z$, and $\delta z\in\mathbb{R}^1$ is an arbitrary scale.

Under Condition C1, the rotational kinematics equivalence in Eq. \eqref{eq72} holds. Substituting this into the Lie group variational framework yields the following critical relationship: 
 \begin{align}
 	&{\mathrm{R}}_{j}\frac{\partial \Delta {\mathrm{R}}_{Fij}}{\partial \mathbf{b}_{Fgi}}{\mathrm{R}}_{i}^\top\boldsymbol{\omega}_{Fi}^L\nonumber\\
 	&=-\Sigma_{k=i}^{j-1}{\mathrm{R}}_{j}\Delta{\mathrm{R}}_{Fk+1j}^\top\mathrm{J}_{Frk}\Delta t{\mathrm{R}}_{i}^\top\boldsymbol{\omega}_{Fi}^L\nonumber\\
 	&\simeq-\Sigma_{k=i}^{j-1}{\mathrm{R}}_{j}\Delta{\mathrm{R}}_{Fk+1j}^\top\mathrm{R}_{k+1}^\top\boldsymbol{\omega}_{Fi}^L\Delta t\nonumber\\
 	&=-\Sigma_{k=i}^{j-1}\Delta{\mathrm{R}}_{Lk+1j}\boldsymbol{\omega}_{Fi}^L\Delta t\nonumber\\
 	&\simeq-\Sigma_{k=i}^{j-1}\Delta{\mathrm{R}}_{Lk+1j}^\top\mathrm{J}_{Lrk}\Delta t\boldsymbol{\omega}_{Fi}^L\nonumber\\
 	&=\frac{\partial \Delta {\mathrm{R}}_{Lij}}{\partial \mathbf{b}_{Lgi}}\boldsymbol{\omega}_{Fi}^L,
 \end{align}\label{eq77}
where the approximate relation $\Delta{\mathrm{R}}_{k + 1j}^\top\mathrm{J}_{rk}\simeq\Delta{\mathrm{R}}_{k + 1j}^\top$ is used. This adjoint invariance under constrained rotation directly implies:
 $\mathrm{J}_{\mathrm{R}_j\mathbf{b}_i}\mathbf{b}_{g+}^{\boldsymbol{\omega}}=\mathbf{0}_{3}$. 

For $\mathrm{J}_{\mathbf{v}_j^\prime\mathbf{b}_i}$, by using Eq. \eqref{eq20_}
$\eqref{eq76}$, it follows that	
\begin{align*}
	&{\mathrm{R}}_{i}\frac{\partial \Delta {\mathbf{v}}_{Fij}}{\partial \mathbf{b}_{Fgi}}{\mathrm{R}}_{i}^\top\boldsymbol{\omega}_{Fi}^L\\
	&=-\Sigma_{k=i}^{j-1}{\mathrm{R}}_{i}\Delta{\mathrm{R}}_{Fik}\mathbf{a}_{Fk}^\wedge\frac{\partial \Delta {\mathrm{R}}_{Fij}}{\partial \mathbf{b}_{Fgi}}\Delta t{\mathrm{R}}_{i}^\top\boldsymbol{\omega}_{Fi}^L\nonumber\\
	&=-\Sigma_{k=i}^{j-1}{\mathrm{R}}_{i}\Delta{\mathrm{R}}_{Fik}{\mathrm{R}}_{k}^\top{\mathrm{R}}_{k}\mathbf{a}_{Fk}^\wedge{\mathrm{R}}_{j}^\top{\mathrm{R}}_{j}\frac{\partial \Delta {\mathrm{R}}_{Fik}}{\partial \mathbf{b}_{Fgi}}{\mathrm{R}}_{i}^\top\boldsymbol{\omega}_{Fi}^L\Delta t\nonumber\\
	&=-\Sigma_{k=i}^{j-1}\Delta{\mathrm{R}}_{Lik}\left({\mathrm{R}}_{k}\mathbf{a}_{Fk}\right)^\wedge{\mathrm{R}}_{k}{\mathrm{R}}_{j}^\top\frac{\partial \Delta {\mathrm{R}}_{Lik}}{\partial \mathbf{b}_{Lgi}}\boldsymbol{\omega}_{Fi}^L\Delta t.\nonumber
\end{align*}
From c2 and c2, we can obtain that $\boldsymbol{\omega}_L\parallel\boldsymbol{\omega}_F^L$ or $\boldsymbol{\omega}_L=\mathbf{0}_{3}$. Therefore, $\frac{\partial \Delta {\mathrm{R}}_{Lik}}{\partial \mathbf{b}_{Lgi}}\boldsymbol{\omega}_{Fi}^L\parallel\boldsymbol{\omega}_{Fi}^L$. In addition, according to c3 and Eq. $\eqref{eq76}$, we can get $\left(\mathrm{R}\mathbf{a}_F\right)^\wedge\boldsymbol{\omega}_F^L=\left(\mathbf{a}_L\right)^\wedge\boldsymbol{\omega}_F^L$. Combining with Eq. $\eqref{eq72}$, we have
\begin{align}
	{\mathrm{R}}_{i}\frac{\partial \Delta {\mathbf{v}}_{Fij}}{\partial \mathbf{b}_{Fgi}}{\mathrm{R}}_{i}^\top\boldsymbol{\omega}_{Fi}^L=&-\Sigma_{k=i}^{j-1}\Delta{\mathrm{R}}_{Lik}\mathbf{a}_{Lk}^\wedge\frac{\partial \Delta {\mathrm{R}}_{Lik}}{\partial \mathbf{b}_{Lgi}}\Delta t\nonumber\\
	=&\frac{\partial \Delta {\mathbf{v}}_{Lij}}{\partial \mathbf{b}_{Lgi}}\boldsymbol{\omega}_{Fi}^L.
\end{align}
According to c2 and c4, we can obtain that
\begin{equation}
	\left[\bar{\mathbf{v}}_{j}^{\prime}\times\right]\frac{\partial \Delta \bar{\mathrm{R}}_{Lij}}{\partial \mathbf{b}_{Lgi}}\boldsymbol{\omega}_{Fi}^L=\mathbf{0}_3,
\end{equation}
Then we have $\mathrm{J}_{{\mathbf{v}_j^\prime\mathbf{b}_i}}\mathbf{b}_{g+}^{\boldsymbol{\omega}}=\mathbf{0}_{3}$. 

For $\mathrm{J}_{\mathbf{p}_j\mathbf{b}_i}$, through a similar process it follows that  
\begin{align}
	{\mathrm{R}}_{i}\frac{\partial \Delta {\mathbf{p}}_{Fij}}{\partial \mathbf{b}_{Fgi}}{\mathrm{R}}_{i}^\top\boldsymbol{\omega}_{Fi}^L=&\frac{\partial \Delta {\mathbf{p}}_{Lij}}{\partial \mathbf{b}_{Lgi}}\boldsymbol{\omega}_{Fi}^L,\\
	\left[\bar{\mathbf{p}}_{j}\times\right]\frac{\partial \Delta \bar{\mathrm{R}}_{Lij}}{\partial \mathbf{b}_{Lgi}}\boldsymbol{\omega}_{Fi}^L=&\mathbf{0}_3,
\end{align}
Then we have $\mathrm{J}_{{\mathbf{p}_j\mathbf{b}_i}}\mathbf{b}_{g+}^{\boldsymbol{\omega}}=\mathbf{0}_{3}$. In summary, we can obtain that $\mathrm{J}_{\mathbf{b}_i}\delta\mathbf{x}_{\mathbf{b}}=\mathrm{J}_{\mathbf{b}_i}\mathbf{b}_{g+}^{\boldsymbol{\omega}}\delta{z}=\mathbf{0}_9$, and $\mathbf{b}_{g+}^{\boldsymbol{\omega}}$ is an unobservable subspace.

\newpage

\vfill

\end{document}